\documentclass[final,1p,times,authoryear]{elsarticle}

\usepackage{amssymb}
\usepackage{amsmath}
\usepackage{amssymb}
\usepackage{amsmath}

\usepackage{xcolor}
\usepackage{float}
\usepackage{algorithm}
\usepackage{algpseudocode}

\usepackage{hyperref}
    \hypersetup{
        colorlinks=true,  
        allcolors=red     
    }

\begin{document}

\begin{frontmatter}

\title{A Hybrid Deep Learning based Carbon Price Forecasting Framework with Structural Breakpoints Detection and Signal Denoising}

\author[a]{Runsheng Ren\fnref{co1}}
\author[a]{Jing Li\fnref{co1}}
\author[a]{Yanxiu Li}
\author[a]{Shixun Huang\corref{cor1}}
\ead{shixunh@uow.edu.au}
\author[a]{Jun Shen}
\author[a]{Wanqing Li}
\author[a]{John Le}
\author[b]{Sheng Wang}

\address[a]{School of Computing and Information Technology, University of Wollongong, Wollongong, Australia}
\address[b]{School of Computer Science, Wuhan University, Wuhan, China}
\cortext[cor1]{Corresponding author.}

\fntext[co1]{These authors contributed equally as co-first authors.}

\begin{abstract}
Accurately forecasting carbon prices is essential for informed energy market decision-making, guiding sustainable energy planning, and supporting effective decarbonization strategies. However, it remains challenging due to structural breaks and high-frequency noise caused by frequent policy interventions and market shocks. Existing studies, including the most recent baseline approaches, have attempted to incorporate breakpoints but often treat denoising and modeling as separate processes and lack systematic evaluation across advanced deep learning architectures, limiting the robustness and the generalization capability. To address these gaps, this paper proposes a comprehensive hybrid framework that integrates structural break detection (Bai–Perron, ICSS, and PELT algorithms), wavelet signal denoising, and three state-of-the-art deep learning models (LSTM, GRU, and TCN). Using European Union Allowance (EUA) spot prices from 2007 to 2024 and exogenous features such as energy prices and policy indicators, the framework constructs univariate and multivariate datasets for comparative evaluation. Experimental results demonstrate that our proposed PELT-WT-TCN achieves the highest prediction accuracy, reducing forecasting errors by 22.35\% in RMSE and 18.63\% in MAE compared to the state-of-the-art baseline model (Breakpoints with Wavelet and LSTM), and by 70.55\% in RMSE and 74.42\% in MAE compared to the original LSTM without decomposition from the same baseline study. These findings underscore the value of integrating structural awareness and multiscale decomposition into deep learning architectures to enhance accuracy and interpretability in carbon price forecasting and other nonstationary financial time series. 

\end{abstract}


\end{frontmatter}

\section{Introduction}

\noindent \textbf{Background}. Against the backdrop of increasingly severe global climate change, carbon markets have become a vital economic tool for controlling carbon emissions. Among them, the European Union Emissions Trading System (EU ETS) is the most mature and influential carbon trading framework in the world~\citep{EuropeanCommission2020}. Within this system, European Union Allowances (EUA), the de facto tradable permits for greenhouse gas emissions, serve as a market indicator. Their prices (i.e., the "EU carbon price")~\citep{KOCH2014676} not only reflect supply and demand dynamics but also convey changes in macroeconomic conditions and policy signals. Therefore, accurately forecasting EUA carbon prices is crucial for policy formulation, enabling hybrid data-driven and physical modeling approaches, supporting intelligent energy management, guiding decarbonization roadmaps, and addressing both energy system and societal impacts.

However, the dynamic nature of the carbon market makes forecasting particularly complex. Unlike traditional financial assets, carbon prices are heavily influenced by frequent policy shocks, such as cap adjustments, energy price volatility, and geopolitical disruptions ~\citep{KOCH2014676}. These external factors often result in structural breaks and high-frequency noise~\citep{lin2022forecasting}, making carbon prices highly nonlinear and nonstationary, which poses serious challenges to existing modeling and forecasting techniques.

\smallskip
\noindent \textbf{Existing Approaches and their Limitations}.
Although numerous studies have attempted to forecast EUA carbon prices ~\citep{huang2021, lin2022forecasting}, most models still rely on linear econometric approaches or fail to fully account for structural changes and noise in the carbon price series. For example, models such as ARIMA~\citep{box1976}, GARCH \citep{bollerslev1986} have been widely applied. More recently, deep learning frameworks, including LSTM~\citep{hochreiter1997}, GRU~\citep{cho2014}, and carbon price specific approaches \citep{Zhang2022} have also emerged. While these models demonstrate strong predictive performance during stable periods, they often suffer from underfitting, lagged responses, and poor generalization when confronted with major institutional changes or macroeconomic disruptions, making them inadequate for real-world forecasting tasks.

While some recent works attempt to address these challenges, important limitations still remain. For instance, \citet{Liu2012} applied wavelet analysis to carbon allowance price dynamics, showing that time and frequency methods can capture structural variations in carbon price series. Building on such approaches, later studies~\citep{lin2022forecasting} have developed hybrid models that integrate structural breakpoint detection (e.g., Bai–Perron and ICSS algorithms) and wavelet based denoising with deep learning architectures such as LSTM, which significantly improved prediction accuracy. However, these studies typically employ only a single deep learning architecture (LSTM) without comparative evaluation of alternative models (e.g., GRU, TCN~\citep{bai2018empirical}), limiting the robustness assessment. On the other hand, breakpoint detection is conducted solely on carbon price series, neglecting macroeconomic or policy driven external features, which limits interpretability. Moreover, most of these methods rely on Bai–Perron or ICSS, which are computationally intensive and less flexible for large-scale datasets. To overcome this, our study adopts the PELT algorithm, which offers linear computational complexity and can detect multiple change points efficiently, making it particularly suitable for incorporating multivariate external factors alongside carbon price series. Further, despite employing wavelet denoising, they do not systematically explore multiscale decomposition, potentially missing important frequency domain dynamics. 

\smallskip
\noindent \textbf{Our goals and Innovations}. To address the aforementioned issues, our proposed framework integrates PELT breakpoints detection, wavelet signal decomposition, comparative evaluation of multiple deep learning models, and the incorporation of multisource external features. This integrated design significantly enhances the robustness, accuracy, and interpretability of carbon price forecasting by:
\begin{itemize}
    \item \textbf{Integrating methods into a unified pipeline}: aligning training data with stable market regimes through breakpoint detection~\citep{Killick2012,bai2003,inclan1994use}, applying wavelet analysis to highlight key dynamics in carbon price data~\citep{Liu2012}, and linking breaks to specific policy or economic events for clearer interpretation~\citep{lin2022forecasting}.  
    \item \textbf{Comparative evaluation of multiple deep learning architectures}: implementing LSTM \citep{hochreiter1997}, GRU~\citep{cho2014}, and TCN~\citep{bai2018empirical} to systematically assess robustness and generalizability across different scenarios.  
    \item \textbf{Incorporating external features}: expanding the feature space with commodity prices, exchange rates, and policy signals to capture macroeconomic and policy-driven influences on carbon price dynamics.  
\end{itemize}


This study also builds upon a real EUA carbon market dataset, incorporating multiple external variables such as coal and natural gas prices~\citep{lin2022forecasting}, policy events~\citep{fan2025cbam}, and broader financial indicators (e.g., exchange rates) that have been considered in recent deep learning approaches \citep{zhao2023ssrn}. A new multifactor enhanced dataset is constructed to improve the feature richness and realism of the training data.

\smallskip
\noindent \textbf{Contributions}. The main contributions of this study are summarized as follows: 
\begin{itemize}
    \item We propose a comprehensive forecasting framework that tightly couples structural break detection, wavelet denoising, and deep learning modeling in a unified pipeline. Unlike prior modular approaches that treat these components independently, we leverage detected breakpoints to guide the segmentation and wavelet decomposition of the time series, enabling the learning models to focus on cleaner, regime consistent data. This integration improves robustness to structural changes and enhances forecasting accuracy under high market volatility, directly addressing the limitations of baseline methods.

    \item We enrich the feature set by incorporating commodity prices, exchange rates, and policy signals, thereby expanding input diversity and enabling the models to better capture the economic and policy-driven dynamics underlying carbon price fluctuations.
    \item We implement and evaluate multiple deep learning models—\textbf{PELT-WT-LSTM} (both univariate and multivariate variants), \textbf{PELT-WT-GRU}, and \textbf{PELT-WT-TCN}—and conduct a systematic comparison against the baseline BP\&ICSS-WT-LSTM. This allows us to assess the effectiveness of alternative architectures beyond traditional LSTM models.
    \item Our experiments demonstrate that incorporating breakpoint aware decomposition significantly improves forecasting accuracy and robustness under volatile market conditions. Our best-performing method, \textbf{PELT-WT-TCN}, achieves a MAE of 1.1855 and an RMSE of 1.5866. This performance indicates an 18.63\% reduction in MAE and a 22.35\% reduction in RMSE respectively, compared to the strongest baseline model, and a 70.55\% reduction in RMSE and a 74.42\% reduction in MAE respectively, compared to the weakest baseline.

\end{itemize}

\smallskip
\noindent \textbf{Structure of the Remainder of this Paper}. Section \ref{sec:RelatedWork}: Related work reviews previous studies and defines our approach, including structural breakpoint detection, time series modeling, and model research; Section \ref{sec:OurMethodology}: Methodology describes the research methods, including breakpoint detection, wavelet denoising, and model architecture design; Section \ref{sec:Experiment}: Experiments present the results along with their implications and limitations;
Section \ref{sec:Conclusion}: Conclusion summarizes this work and outlines future directions.

\section{Related Work}
\label{sec:RelatedWork}

\noindent \textbf{Signal Decomposition and Deep Learning Models.}  
Several recent studies~\citep{zhang2024,zhang2023,yue2023} have explored hybrid models that combine signal decomposition techniques with machine learning or deep learning architectures. For instance, a deep learning based framework including multisource features was proposed to improve prediction performance in carbon markets \citep{Zhang2022}. A multifrequency combined model was also introduced to capture both long-term and short-term patterns in price dynamics \citep{Duan2025}. In addition, quantile regression with feature selection has been applied to improve accuracy under uncertain market conditions \citep{pang2023carbonpriceforecastingquantile}. These methods generally aim to denoise input data and capture nonlinear structures, often using tools such as EMD \citep{lorentz1938limitations}, CEEMDAN \citep{torres2011ceemdan}, or wavelet transforms \citep{Mallat1989}. While such approaches have shown promise in improving accuracy, most do not explicitly address the role of structural breakpoints, or only incorporate them in a limited fashion \citep{Zhang2022,Duan2025}. Furthermore, few studies \citep{torres2011ceemdan,shahid2020,Zhang2022}
conduct systematic comparisons of deep learning models under standardized conditions. It is therefore difficult to assess their generalizability, robustness and computational cost.

\noindent \textbf{Model Comparisons and Advanced Neural Architectures.}  
Although the application of deep learning in carbon price forecasting has rapidly expanded in recent years, systematic comparisons of different models under standardized conditions remain scarce. This lack of comparative studies limits a comprehensive evaluation of these methods in terms of generalization ability, robustness, and computational cost. To address this gap, several studies have explored the issue from various perspectives. One study combined GARCH and LSTM models to handle the high volatility of the EU carbon market, demonstrating the potential of hybrid modeling in multiscale forecasting~\citep{huang2021}. Another study incorporated attention mechanisms into a BiLSTM network to improve its ability to model degradation trends, thereby highlighting how structural modifications can impact model performance~\citep{guo2023}. Ensemble deep learning methods based on high-frequency trading data were evaluated to emphasize the coupling between data granularity and model compatibility~\citep{zhao2023ssrn}. Additionally, prediction accuracy has been examined from the perspectives of model integration~\citep{chen2021} and regional heterogeneity~\citep{zhao2023}. An integrated CEEMDAN and TCN-LSTM framework has also been proposed to enhance adaptability and accuracy under nonlinear volatility~\citep{cai2025}.

\noindent \textbf{Model Interpretability and Performance Optimization.}  
In the area of model interpretability and decision support, SHAP and LIME techniques have been applied to improve the transparency and practical applicability of AI models in carbon market forecasting~\citep{olawumi2025,lei2024}. Comparative validation has been conducted within an SSA-LSTM framework to provide empirical support for methodological evaluation~\citep{wang2022carbon}. Furthermore, LightGBM combined with Bayesian optimization has been used to assess error convergence across multiparameter spaces, emphasizing the importance of structural tuning and performance optimization~\citep{deng2024}. Collectively, these studies underscore the critical need to establish standardized evaluation benchmarks and cross model performance comparison systems to support the sustainable development of intelligent forecasting in carbon markets.

\noindent \textbf{Breakpoint Detection and Carbon Price Forecasting.}  
The baseline paper \citep{lin2022forecasting} proposed a forecasting framework that integrates structural breakpoint detection with deep learning and highlights the value of capturing policy changes to improve the prediction of carbon prices. While this provides a useful foundation for hybrid modeling approaches that account for market dynamics, the structural breakpoint detection methods adopted are relatively dated and more suitable for identifying linear structural changes. This limits their ability to capture more complex or nonlinear regime shifts that are often present in volatile markets such as carbon trading. Building on this line of research, our study optimizes the breakpoint detection method together with wavelet based signal decomposition. In addition, we systematically evaluate several deep learning models using the preprocessed data (univariate, multivariate), including standard LSTM without structural features, univariate LSTM\citep{hochreiter1997}, multivariate LSTM \citep{karim2019multivariate}, GRU\citep{shahid2020}, and TCN\citep{chang2025}. All models are trained and tested under consistent conditions to enable a fair comparison across different input configurations.

\section{Our Methodology}
\label{sec:OurMethodology}

\begin{figure}[H]
  \centering
  \includegraphics[width=10cm]{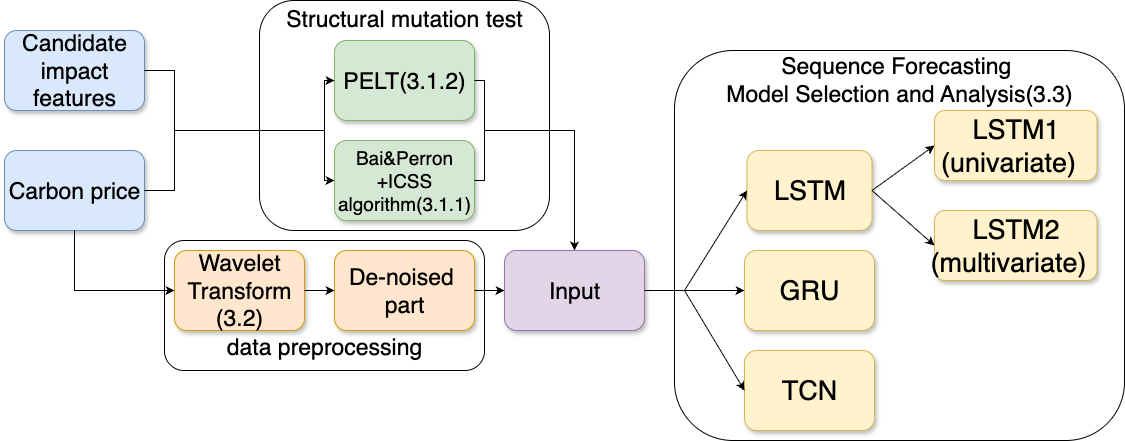}
  \caption{The flowchart of the proposed framework. This framework outlines a carbon price forecasting process combining structural break detection, wavelet denoising, and sequence models (LSTM, GRU, TCN) for univariate and multivariate analysis.}
  \label{fig:FW}
\end{figure}

This study formulates the task of carbon price forecasting as a multivariate, one-step regression problem, aiming to predict the carbon allowance price at the next time point based on historical carbon price data. As illustrated in Figure~\ref{fig:FW}, the overall prediction framework encompasses key stages such as data construction, preprocessing, feature extraction, and model forecasting, forming a systematic and generalizable approach to carbon price prediction. First, the study constructs the raw dataset based on actual carbon market data, and in the preprocessing stage, the carbon price series undergoes structural break detection and wavelet denoising to generate input features that are more stable and trend oriented. Specifically, Sections \ref{subsec:BP} and \ref{subsec:PELT} present the theoretical foundations and implementation of the Bai-Perron+ICSS and PELT algorithms, which identify potential breakpoints in the time series from different perspectives to enhance the robustness and interpretability of subsequent modeling. Section \ref{subsec:wavelet} explains the application of wavelet transform to separate different frequency components, remove high-frequency noise, and retain the main trend signal, thereby improving the quality of extracted features. Based on this, the denoised carbon price series is processed with time lags to construct input variables that reflect temporal dependencies, which are then integrated with structural break information to form the final modeling feature set. Subsequently, Section~\ref{subsec:Sequence} introduces typical deep learning models such as LSTM, GRU, and TCN, detailing their structural principles and suitability for nonlinear time series prediction, and demonstrating their respective advantages for carbon price forecasting tasks. Furthermore, Section~\ref{subsec:Comparison} compares the detection performance of different structural break detection methods, discusses the rationale for model selection, andvisualizes the overall prediction process and the connections between each stage. Through the integration of these steps, the proposed framework effectively achieves the transformation from raw data to structured and denoised features, combines key breakpoints identification with deep learning model training, and ultimately outputs more accurate and interpretable carbon price predictions.

\subsection{Underlying Models}
\subsubsection{Bai Perron + ICSS}
\label{subsec:BP}

\smallskip
\noindent \textbf{Bai-Perron Structural Break Test}.  
The Bai–Perron test~\citep{bai2003} is a classical method for detecting multiple structural breakpoints in time series data.  
It partitions the series $y_t$ into $m+1$ regimes by minimizing the residual sum of squares (RSS). Formally, the model is expressed as a piecewise linear regression:  
\begin{equation}
y_t = x_t^\top \beta_j + u_t, \quad t = \tau_{j-1}+1,\dots,\tau_j, \; j=1,\dots,m+1
\label{eq:A}
\end{equation}
where
\begin{itemize}
  \item $y_t$: the dependent variable (e.g., carbon price) at time $t$.
  \item $x_t$: a vector of explanatory variables (regressors) at time $t$.
  \item $\beta_j$: the coefficient vector specific to regime $j$.
  \item $u_t$: the error term at time $t$.
  \item $\tau_j$: the $j$-th structural break point.
  \item $m$: the number of structural breaks in the series.
\end{itemize}

The breakpoints $\{\tau_1, \dots, \tau_m\}$ are estimated by minimizing the overall residual sum of squares:
\begin{equation}
S_T = \min_{\{\tau_1, \dots, \tau_m\}}
\sum_{j=1}^{m+1} \sum_{t=\tau_{j-1}+1}^{\tau_j} 
\left( y_t - x_t^\top \hat{\beta}_j \right)^2
\label{eq:B}
\end{equation}

To evaluate significance, the $SupF$ test statistic is constructed:
\begin{equation}
\mathrm{Sup}F = 
\max_{k=\lceil 1+\eta T \rceil}^{\lfloor T-\eta T \rfloor}
\frac{S_0 - S_k}{\widehat{\sigma}^2}
\label{eq:C}
\end{equation}
where $S_0$ is the RSS under the null hypothesis (no break), $S_k$ is the RSS under the alternative (break at $k$), and $\hat{\sigma}^2$ is the variance estimate.

\smallskip
\noindent \textbf{ICSS Algorithm}.  
The Iterative Cumulative Sum of Squares (ICSS) algorithm~\citep{inclan1994use} detects structural changes in the variance of the same time series $y_t$.  
The observed series can be expressed as:
\begin{equation}
y_t = \mu + e_t, \quad t = 1, 2, \ldots, T
\end{equation}
where $\mu$ is the mean and $e_t$ is the residual term.  

To test for variance shifts, the cumulative variance statistic is defined as:
\begin{equation}
C_k = \sum_{i=1}^k z_i^2, \quad 
z_t = \frac{y_t - \bar{y}_{1:t}}{\sqrt{\tfrac{1}{t}+\tfrac{1}{T}} \, S_y}, \quad t=1,\dots,T
\end{equation}
where $\bar{y}_{1:t}$ is the mean of the first $t$ observations and $S_y$ is the sample standard deviation.

\noindent The normalized deviation is:
\begin{equation}
D_k = \frac{C_k}{C_T} - \frac{k}{T}, \quad D_0 = D_T = 0,
\end{equation}
and the test statistic is:
\begin{equation}
IT = \sup \left( \sqrt{ \tfrac{T}{2} } |D_k| \right).
\end{equation}

\noindent If $IT$ exceeds a critical threshold, a variance break is identified.  

\smallskip
\noindent \textbf{Unified Output}.  
Both Bai–Perron (mean/trend changes) and ICSS (variance shifts) produce breakpoints $\{\tau_j\}$, which are encoded into regime labels $r_t$. These regime labels serve as structured inputs for subsequent forecasting models.

\subsubsection{PELT}
\label{subsec:PELT}

\smallskip
\noindent \textbf{Multiple breakpoints Detection}.  
The Pruned Exact Linear Time (PELT) algorithm~\citep{Killick2012} detects multiple breakpoints in $y_t$ by minimizing the cost function:
\begin{equation}
\sum_{i=1}^{m+1} C(y_{\tau_{i-1}+1:\tau_i}) + \beta m,
\label{eq:pelt-cost}
\end{equation}
where $C(\cdot)$ is the segment cost (e.g., negative log-likelihood), $\beta$ is the penalty, and $\{\tau_1,\dots,\tau_m\}$ are breakpoints.

\smallskip
\noindent \textbf{Recursive Formulation}.  
Let $F(t)$ denote the optimal segmentation cost up to time $t$:  
\begin{equation}
F(t) = \min_{s<t} \Bigl\{ F(s) + C(y_{s+1:t}) + \beta \Bigr\}.
\label{eq:F}
\end{equation}
This recursion prunes suboptimal candidates, achieving $\mathcal{O}(n)$ complexity in both time and space.  

\smallskip
\noindent \textbf{Unified Output}.  
The resulting breakpoints $\{\tau_j\}$ are also encoded into regime labels $r_t$, ensuring consistency with Bai–Perron and ICSS. This unified representation allows fair comparison of forecasting models under different breakpoint detection strategies.


\subsection{Wavelet Transform}
\label{subsec:wavelet}
\begin{figure}[H]\centering
\includegraphics[width=9cm]{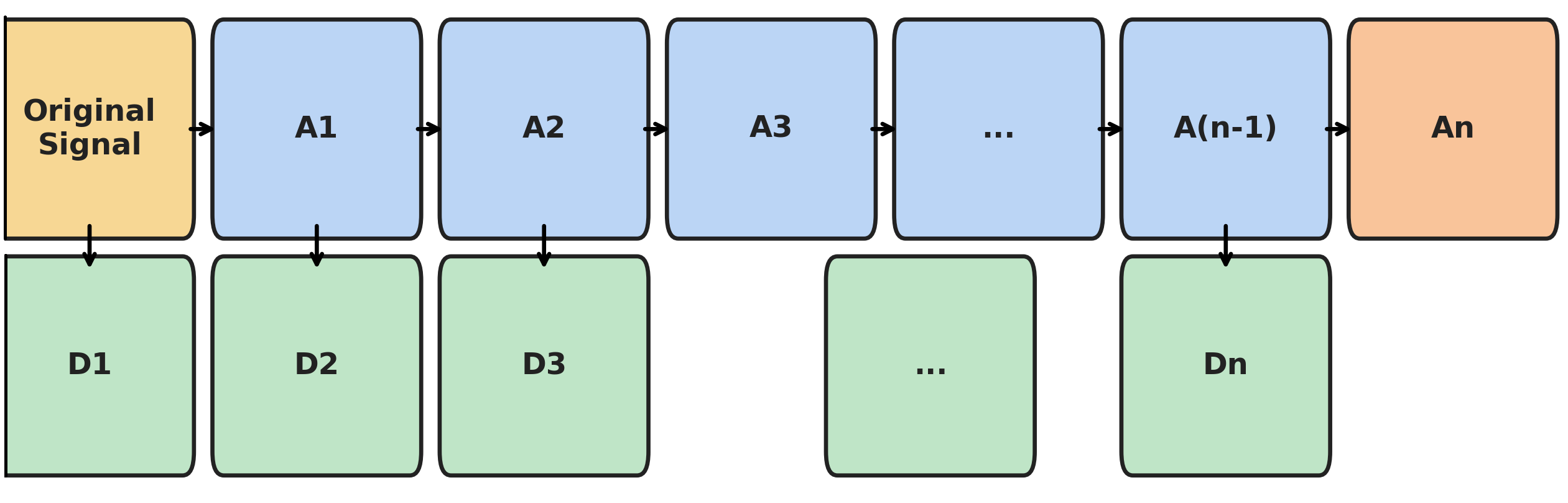}
  \caption{Flowchart of wavelet‐transform decomposition.}
  \label{fig:wt-decomp}
\end{figure}

Wavelet transform (WT) is a time–frequency analysis tool that projects a
signal~$f(t)$ onto a family of dilated and translated basis
functions, thereby revealing localised spectral information.
Owing to its multiresolution capability, WT has become a standard
procedure for decomposing nonstationary series into an
\emph{approximation component} (low frequency, denoised) and a
\emph{detail component} (high-frequency)~\citep{Mallat1989}. 
Figure~\ref{fig:wt-decomp} illustrates the multilevel wavelet decomposition pipeline adopted in this study, where the original signal is first decomposed into an approximation component (A1) and a detail component (D1). The approximation component is then recursively decomposed into further approximation (A2, A3, …) and detail components (D2, D3, …) at subsequent levels.

\paragraph{Nested subspaces}
\begin{equation}
V_j \subset V_{j+1}, \quad j \in \mathbb{Z}
\end{equation}
\noindent where
\begin{itemize}
  \item $V_j$: approximation subspace at resolution $2^j$;
  \item $j$: scale index.
\end{itemize}

\paragraph{Limit properties}
\begin{equation}
\overline{\bigcup_{j\in\mathbb{Z}}}V_j = L^2(\mathbb{R}),
\qquad 
\bigcap_{j\in\mathbb{Z}} V_j = \{0\}
\end{equation}
\noindent where
\begin{itemize}
  \item $L^2(\mathbb{R})$: space of square-integrable functions;
  \item $V_j$: approximation subspace.
\end{itemize}

\paragraph{Scaling function and orthonormal basis}
\begin{equation}
\varphi_{j,n}(x) = 2^{j/2}\,\varphi(2^j x - n),
\qquad
\{\varphi_{j,n}(x)\}_{n\in\mathbb{Z}} 
\end{equation}
\noindent where
\begin{itemize}
  \item $\varphi(x)$: scaling function (father wavelet);
  \item $\varphi_{j,n}(x)$: scaling basis at scale $j$ and translation $n$;
  \item $j$: scale index (dilation);
  \item $n$: translation index (shift).
\end{itemize}

\paragraph{Orthogonal expansion}  
\begin{equation}
A_j f(x) = \sum_{n\in\mathbb{Z}} \langle f, \varphi_{j,n}\rangle \varphi_{j,n}(x)
\end{equation}
\noindent where
\begin{itemize}
  \item $A_j f$: projection of $f$ onto subspace $V_j$ (approximation);
  \item $\langle f, \varphi_{j,n}\rangle$: approximation coefficient;
  \item $\varphi_{j,n}(x)$: scaling basis function.
\end{itemize}

\paragraph{Convolution + downsampling (pyramid algorithm)}  
\begin{equation}
\langle f, \varphi_{j,n}\rangle 
= \sum_{k\in\mathbb{Z}} \overline{h}(2n-k)\,\langle f, \varphi_{j-1,k}\rangle
\end{equation}
\noindent where
\begin{itemize}
  \item $h(n)$: low-pass filter associated with the scaling function;
  \item $f$: original signal;
  \item $\varphi_{j-1,k}$: scaling basis at the previous level.
\end{itemize}

\paragraph{Frequency-domain definition of the wavelet}  
\begin{equation}
\hat{\psi}(\omega) = G\!\left(\tfrac{\omega}{2}\right)\,\hat{\varphi}\!\left(\tfrac{\omega}{2}\right),
\qquad 
G(\omega) = e^{-i\omega}\,\overline{H(\omega+\pi)}
\end{equation}
\noindent where
\begin{itemize}
  \item $\psi(x)$: wavelet function (mother wavelet);
  \item $\hat{\psi}(\omega)$: Fourier transform of the wavelet function;
  \item $\varphi(x)$: scaling function;
  \item $H(\omega)$: low-pass filter in the Fourier domain;
  \item $G(\omega)$: high-pass filter in the Fourier domain.
\end{itemize}

\paragraph{Detail signal expansion}  
\begin{equation}
D_j f = \Big(\langle f, \psi_{j,n}\rangle\Big)_{n\in\mathbb{Z}}
\end{equation}
\noindent where
\begin{itemize}
  \item $D_j f$: detail component at scale $j$;
  \item $\psi_{j,n}(x)$: wavelet basis at scale $j$, translation $n$;
  \item $\langle f,\psi_{j,n}\rangle$: detail coefficient.
\end{itemize}

\paragraph{Complete wavelet decomposition}  
\begin{equation}
f = A_J f + \sum_{j=J}^{\infty} D_j f
\end{equation}
\noindent where
\begin{itemize}
  \item $f$: original signal;
  \item $A_J f$: approximation component at the coarsest scale $J$;
  \item $D_j f$: detail components at scales $j \geq J$.
\end{itemize}

\subsection{Sequence Forecasting Models}
\subsubsection{Long Short-Term Memory (LSTM)}
\label{subsec:Sequence}

\begin{figure}[H]\centering
\includegraphics[width=9cm]{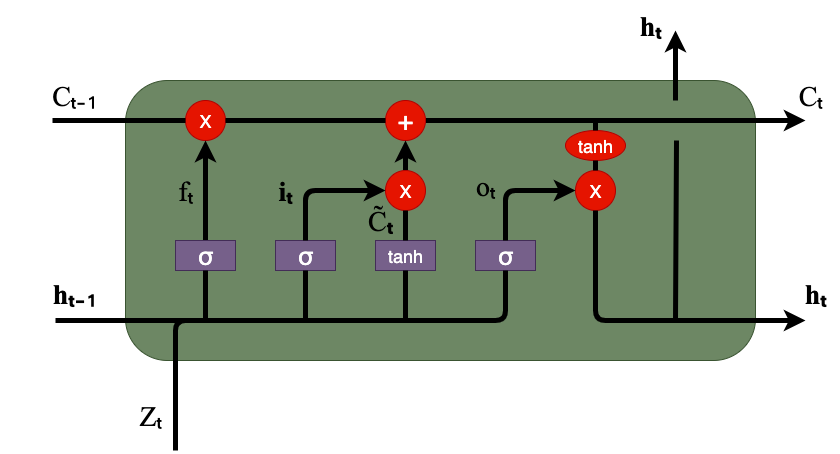}
\caption{Architecture of a peephole-free LSTM cell with the unified input $z_t=[\tilde{y}_t, u_t, e_t]$.}
\label{fig:lstm-unit}
\end{figure}

\noindent \textbf{Long Short‐Term Memory (LSTM)}. Long short-term memory (\textbf{LSTM}) network is a gated variant of the
recurrent neural network (RNN) originally proposed by~\citet{hochreiter1997}.  
By introducing gating mechanisms, LSTM alleviates the
\emph{vanishing‐gradient} problem that plagues conventional RNNs,
thereby retaining long range dependencies in time series.  
It has become the de facto backbone for sequential modelling across
finance, speech and language domains.  
Figure~\ref{fig:lstm-unit} illustrates the architecture of the peephole-free LSTM cell employed in our study, which takes the unified input $z_t = [\tilde{y}_t, u_t, e_t]$.

\vspace{.5em}
\noindent

\noindent Following Sections~3.1 and~3.2, we construct a unified input vector
\begin{equation*}
z_t =
\begin{bmatrix}
\tilde y_t \\
u_t \\
e_t
\end{bmatrix}
\in \mathbb{R}^d,
\end{equation*}
where $\tilde y_t$ denotes the wavelet-denoised carbon price (from Sec.~3.2), 
$u_t \in \mathbb{R}^p$ represents exogenous variables, 
and $e_t \in \mathbb{R}^{m+1}$ is the one-hot encoded regime label from breakpoint detection (Sec.~3.1). 
At each time step $t$, the LSTM cell receives this input $z_t$, together with the previous hidden state $h_{t-1} \in \mathbb{R}^h$ and the previous cell state $C_{t-1} \in \mathbb{R}^h$. 
Three gates---forget, input, and output---jointly regulate the information flow: 

\begin{align}
f_t &= \sigma(W_f z_t + U_f h_{t-1} + b_f), \tag{18}\\
i_t &= \sigma(W_i z_t + U_i h_{t-1} + b_i), \tag{19}\\
\tilde C_t &= \tanh(W_a z_t + U_a h_{t-1} + b_a), \tag{20}
\end{align}

where
\begin{itemize}
    \item $f_t$: the forget gate activation at time $t$, controlling how much of the previous memory to retain.
    \item $i_t$: the input gate activation at time $t$, controlling how much new information to write to the cell state.
    \item $\tilde C_t$: the candidate cell state (candidate activation), computed using a $\tanh$ activation. 
    \item $z_t$: the unified input vector, consisting of the wavelet-denoised carbon price $\tilde y_t$, the exogenous variables $u_t$, and the one-hot regime label $e_t$.
    \item $h_{t-1}$: the hidden state from the previous time step.
    \item $W, U$: weight matrices for input and hidden state, respectively, specific to each gate.
    \item $b$: the bias term corresponding to each gate.
    \item $\sigma(\cdot)$: the logistic sigmoid activation function, producing values in the range $(0,1)$.
    \item $\tanh(\cdot)$: the hyperbolic tangent function, producing values in the range $(-1,1)$.
\end{itemize}

\paragraph{Cell-state update}  
The forget gate $f_t \odot c_{t-1}$ determines how much historical information to retain, 
while $i_t \odot \tilde C_t$ writes the current candidate content into the memory cell. 
The cell state is updated as
\begin{equation}
C_t = \sigma(f_t \odot C_{t-1} + i_t \odot \tilde C_t), \tag{21}
\end{equation}
where
\begin{itemize}
    \item $C_t$: the cell state at time $t$, representing the long-term memory of the LSTM unit.
    \item $C_{t-1}$: the previous cell state (from time $t-1$).
    \item $f_t$: the forget gate value, controlling how much of $c_{t-1}$ is retained.
    \item $i_t$: the input gate value, controlling how much new information is written into the cell.
    \item $\tilde C_t$: the candidate cell state (or activation), proposed as new content.
    \item $\odot$: element-wise (Hadamard) multiplication.
\end{itemize}

\paragraph{Hidden-state output}  
Finally, the output gate selects a portion of the updated cell state as the hidden representation:
\begin{align}
o_t &= \sigma(W_o z_t + U_o h_{t-1} + b_o), \tag{22}\\
h_t &= o_t \odot \tanh(C_t), \tag{23}
\end{align}
where
\begin{itemize}
    \item $o_t$: the output gate at time $t$, determining how much of the internal cell state is exposed to the output.
    \item $W_o, U_o$: weight matrices for the current input $z_t$ and previous hidden state $h_{t-1}$ in the output gate.
    \item $b_o$: the bias vector for the output gate.
    \item $h_t$: the hidden state (i.e., the actual output of the LSTM cell at time $t$).
    \item $C_t$: the updated internal cell state.
    \item $\odot$: element-wise (Hadamard) multiplication.
    \item $\tanh(C_t)$: the nonlinear transformation of the cell state to produce the output.
\end{itemize}

\subsubsection{Gated Recurrent Unit (GRU)}

The Gated Recurrent Unit (GRU) is a streamlined alternative to the LSTM, 
retaining only two gates---update and reset---and discarding the explicit memory cell. 
Despite its compactness, GRU still mitigates vanishing gradients and has shown competitive 
performance in sequence modelling tasks, such as pandemic trajectory forecasting (Shahid et al., 2020). 

At time step $t$, given the unified input vector $z_t \in \mathbb{R}^d$ (defined in Sec.~3.3.1) 
and the previous hidden state $h_{t-1} \in \mathbb{R}^h$, the gates are computed as
\begin{align}
z_t^{(g)} &= \sigma(W_z z_t + U_z h_{t-1} + b_z), \tag{24}\\
\rho_t &= \sigma(W_\rho z_t + U_\rho h_{t-1} + b_\rho), \tag{25}
\end{align}
where
\begin{itemize}
    \item $z_t^{(g)}$: the update gate at time $t$, controlling how much of the past hidden state is preserved.
    \item $\rho_t$: the reset gate at time $t$, determining how much past information to forget.
    \item $z_t$: the unified input vector at time $t$, consisting of $\tilde y_t$, $u_t$, and $e_t$.
    \item $h_{t-1}$: the hidden state from the previous time step.
    \item $W_z, W_\rho$: weight matrices for input-to-gate connections.
    \item $U_z, U_\rho$: weight matrices for hidden-to-gate connections.
    \item $b_z, b_\rho$: bias vectors for the two gates.
    \item $\sigma(\cdot)$: the logistic sigmoid activation function.
\end{itemize}


\textit{Candidate activation.} The reset gate modulates how much past information enters the candidate activation:
\begin{equation}
\tilde h_t = \tanh(W z_t + U (\rho_t \odot h_{t-1}) + b_h), \tag{26}
\end{equation}
where
\begin{itemize}
    \item $\tilde h_t$: the candidate activation at time $t$, representing the new content to be potentially added to the hidden state.
    \item $W, U$: weight matrices for input-to-hidden and hidden-to-hidden connections.
    \item $z_t$: the unified input vector at time $t$, consisting of $\tilde y_t$, $u_t$, and $e_t$.
    \item $h_{t-1}$: the previous hidden state.
    \item $\rho_t$: the reset gate at time $t$, controlling the contribution of past memory.
    \item $b_h$: bias vector for the candidate activation.
    \item $\odot$: element-wise (Hadamard) multiplication.
    \item $\tanh(\cdot)$: hyperbolic tangent activation function.
\end{itemize}


\textit{Hidden-state interpolation.} Finally, the update gate linearly interpolates between the previous state and the candidate activation:
\begin{equation}
h_t = (1 - z_t^{(g)}) \odot h_{t-1} + z_t^{(g)} \odot \tilde h_t, \tag{27}
\end{equation}
where
\begin{itemize}
    \item $h_t$: the final hidden state (and output) of the GRU cell at time $t$.
    \item $h_{t-1}$: the previous hidden state from time $t-1$.
    \item $z_t^{(g)}$: the update gate, determining how much of the new candidate state $\tilde h_t$ is used.
    \item $\tilde h_t$: the candidate activation computed in Eq.~(26).
    \item $(1 - z_t^{(g)}) \odot h_{t-1}$: the retained part of the previous hidden state.
    \item $z_t^{(g)} \odot \tilde h_t$: the newly added information.
    \item $\odot$: element-wise (Hadamard) multiplication.
\end{itemize}

\noindent
Intuitively, $z_t^{(g)}$ controls the trade-off between retaining past information and overwriting it with new content, while $\rho_t$ determines how much historical context contributes to the candidate activation. 
Owing to this gated interpolation, GRU achieves comparable accuracy to LSTM with fewer parameters, 
which explains its adoption in recent time series studies.


\subsubsection{Temporal Convolutional Network (TCN)}

\begin{figure}[H]
    \centering
    \includegraphics[width=6cm]{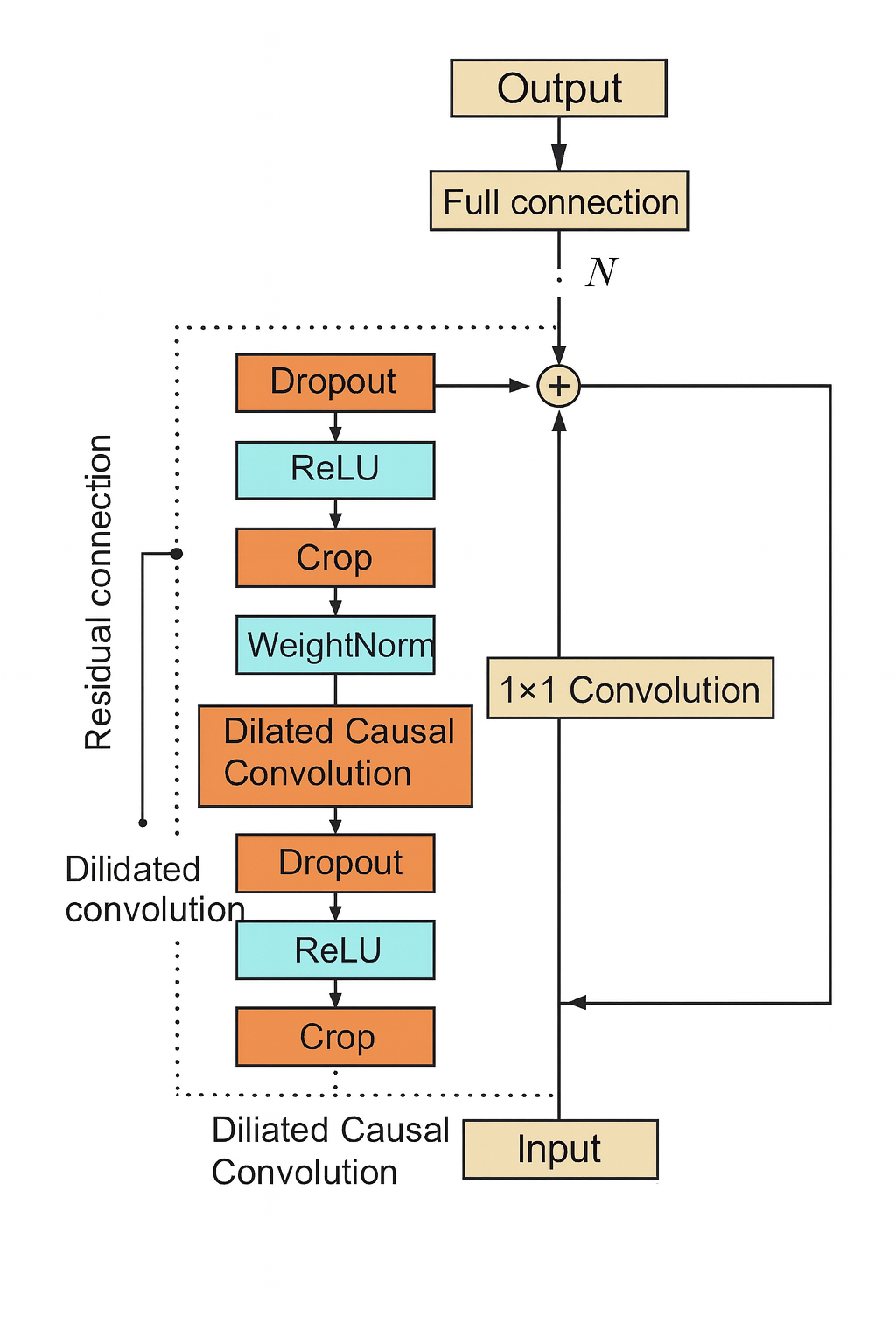}
    \caption{The structure of the TCN model.}
    \label{fig:tcn_structure}
\end{figure}

\noindent \textbf{Temporal Convolutional Network (TCN).} 
The temporal convolutional network (TCN) adopted in this work, originally introduced for grating nanomeasurement by \cite{chang2025}, comprises an \emph{input layer}, $N$ stacked \emph{residual blocks}, a \emph{fully connected} projection, and an \emph{output node}.

At time step $t$, the TCN receives a windowed input matrix
\[
X_t = [z_{t-T+1}, \ldots, z_t] \in \mathbb{R}^{T \times d},
\]
where each $z_\tau$ is the unified input vector defined in Sec.~3.3.1, consisting of the wavelet-denoised carbon price $\tilde y_\tau$, the exogenous variables $u_\tau$, and the regime one-hot label $e_\tau$. 
This design enables the TCN to capture dependencies across a receptive field of length $T$, in contrast to recurrent models which process one time step at a time. 
The overall dataflow is illustrated in Figure~\ref{fig:tcn_structure}~\citep{chang2025}. 

Each residual block performs:
\begin{enumerate}
  \item \textbf{Dilated causal convolution} (twice): a dilation factor
        $d$ enlarges the receptive field exponentially while preserving
        temporal causality;
  \item \textbf{Weight normalisation} (two layers): stabilises activation
        scale and accelerates convergence;
  \item \textbf{Crop (shear)}: aligns sequence length after dilation and
        clips gradients to avert numerical overflow;
  \item \textbf{ReLU activation}: injects nonlinearity after each
        convolution;
  \item \textbf{Dropout}: randomly masks neurons to curb overfitting.
\end{enumerate}

An additional $1{\times}1$ convolution on the shortcut branch matches
channel dimensions before summation. Finally, a dense layer maps the
hidden representation to a scalar prediction delivered by the output
node. This cascade allows the TCN to model both short-range and long-range dependencies efficiently and stably.

\subsection{Comparison of related models}
\label{subsec:Comparison}

\noindent \textbf{Comparison of Structural Mutation Tests}.
The multiple structural mutation test method proposed by~\citet{bai2003} is used to analyze carbon market time series with significant institutional changes. The core advantage of this method is that it can detect structural mutation points at multiple unknown locations simultaneously, and allows different regression models to be applied in each segment. This offers a powerful tool for revealing the dynamic relationship between carbon prices and macroeconomic or energy policy regimes.

Although the Bai–Perron method performs well in detecting institutional shifts, it relies on segmented Ordinary Least Squares (OLS) regression, which imposes several limitations. First, it assumes linearity within each segment, making it less suitable for capturing nonlinear or jump type dynamics. Second, the algorithm has high computational complexity, particularly when the sample size is large or when many breakpoints exist.

In contrast, the ICSS (Iterative Cumulative Sum of Squares) method~\citep{inclan1994use} is specifically designed to detect variance changes in time series. It does not require parametric model specification within segments and is particularly useful for identifying abrupt shifts in volatility. However, ICSS assumes the mean of the series is constant and is more sensitive to outliers or serial correlation in the residuals, which may affect its accuracy. 

Compared with both BP and ICSS, the PELT (Pruned Exact Linear Time) method~\citep{Killick2012} is better suited for large-scale and high-frequency time series. It features a linear time complexity $\mathcal{O}(n)$ and employs pruning strategies to significantly reduce redundant search paths, thus maintaining high computational efficiency while preserving detection accuracy. PELT was chosen because breakpoints were measured for each column of data. 

\noindent \textbf{Comparison of Sequence Forecasting Models.}\label{sec:tcn}
Carbon market prices are influenced by multiple factors such as energy prices, policy interventions, supply and demand fluctuations, and seasonality. This results in significant nonlinearity and long-term and short-term dependencies in the price series. Effective forecasting requires models that can capture such complex temporal dynamics.
The Gated Recurrent Unit (GRU) is a recurrent neural network variant designed to retain long-term dependencies via update and reset gates, which helps mitigate the vanishing gradient problem~\citep{shahid2020}. 
In contrast, the Temporal Convolutional Network (TCN) employs dilated causal convolutions and residual connections to model long range dependencies with high efficiency and stability~\citep{bai2018empirical}. 
While GRU is well-suited for sequential processing, its inherent sequential nature limits training speed. TCN, by contrast, supports parallel computation and demonstrates faster convergence, making it advantageous for large scale or multistep forecasting tasks~\citep{bai2018empirical}~\citep{lim2021tft}.

Empirical studies have shown that TCN can outperform GRU
in many time series forecasting benchmarks, particularly in energy and financial domains~\citep{lim2021tft}. As carbon prices exhibit both market and policy driven volatility, a comparative evaluation of GRU and TCN provides valuable guidance in selecting robust modeling architectures for practical forecasting applications.

\section{Experiment}
\label{sec:Experiment}

This section presents a comprehensive experimental analysis structured into four main parts. Section~\ref{sec:expdataset} introduces the dataset, outlines its statistical characteristics, and identifies structural breakpoints that signal major shifts in carbon pricing patterns. It also explains the data preprocessing technique using Wavelet Transform (WT) for denoising. Section~\ref{sec:expdlmodel} compares the performance of deep learning models, i.e., BP\&ICSS-WT-LSTM, PELT-WT-LSTM, PELT-WT-GRU, and PELT-WT-TCN, under various configurations, highlighting their predictive behaviors. Section~\ref{sec:expperformance} provides a quantitative performance evaluation using metrics such as MAE, RMSE, MAPE and R². Section~\ref{sec:expscalabilityresidual} further analyzes the scalability of the models and the distribution of residuals and errors to evaluate models' robustness and generalizability. 

\subsection{Dataset Analysis and Preprocessing}\label{sec:expdataset}
\begin{figure}[H]
  \centering
  \includegraphics[width=8cm]{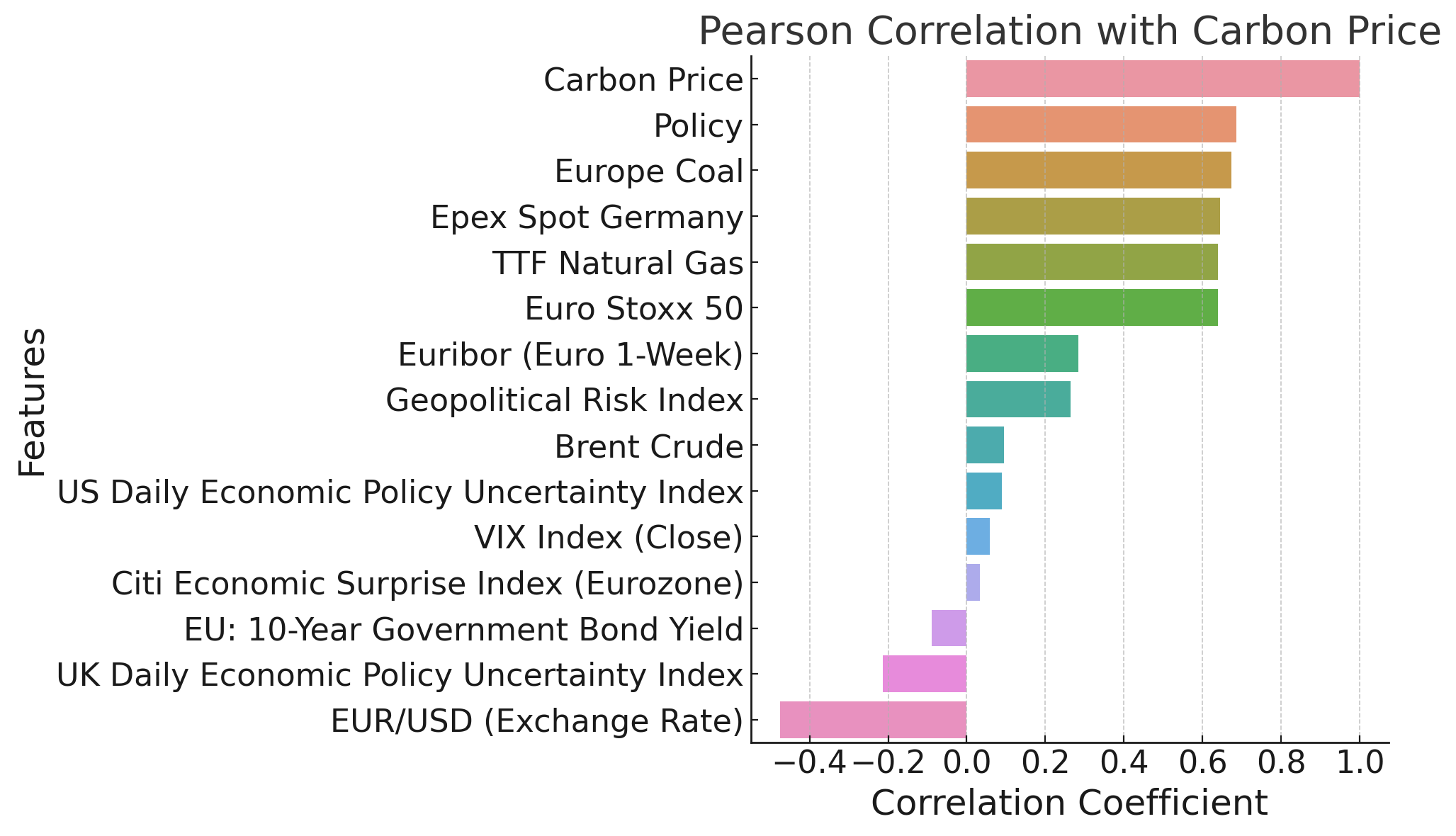}
  \caption{Key drivers of carbon price}
  \label{fig:keydrivers}
\end{figure}

\noindent \textbf{Data Description}.
The daily spot trading price data of EU Allowances (EUA) from September 10, 2007, to June 4, 2024, totaling 6,113 samples, was used for the empirical analysis in this study. This dataset was obtained from the financial market data and information service platform (investing.com). The primary reason for selecting this dataset is its ability to capture the dynamic evolution of the EU carbon trading market across multiple policy phases~\citep{EuropeanCommission2024}. Moreover, the extended time span provides a richer set of training samples for machine learning models, which helps enhance their predictive performance and generalization capability.
\noindent \textbf{Key Impact Feature Analysis}.
Figure~\ref{fig:keydrivers} presents the Pearson correlation coefficients between carbon prices and selected explanatory features.  
The results indicate that policy indicators, coal and natural gas prices, as well as stock market indices (e.g., Euro Stoxx 50) are strongly and positively correlated with carbon prices, highlighting their role as major drivers.  
In contrast, features such as the EUR/USD exchange rate and economic policy uncertainty indices show weaker or even negative correlations.  
These findings suggest that energy market fundamentals and policy dynamics exert the strongest influence on carbon price movements, whereas macroeconomic and financial uncertainty indicators play a more limited role.  
The correlation analysis provides a basis for feature selection in subsequent forecasting models.

\begin{figure}[H]
  \centering
  \includegraphics[width=8cm]{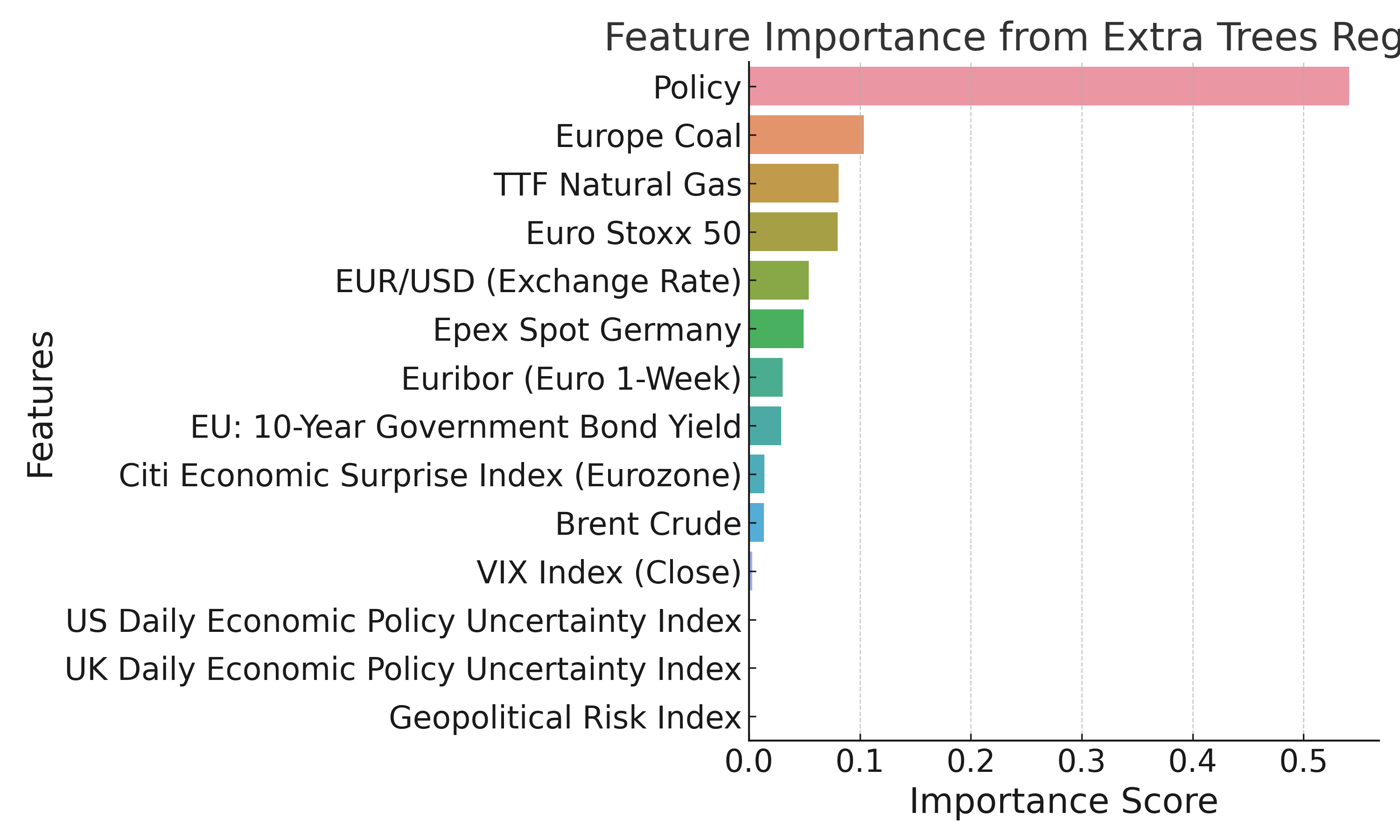}
  \caption{Relative importance of input features for carbon price prediction}
  \label{fig:realimport}
\end{figure}
This study aims to screen multiple external features to identify the key drivers most closely associated with carbon price fluctuations. Following the approach of ~\cite{Geurts2006}, we employed the Extremely Randomized Trees (ET) method to evaluate the importance of input features. As shown in Figure~\ref{fig:realimport}, the results of the ET evaluation indicate that the policy features have the highest importance score, exceeding 0.5, which is significantly higher than the other characteristics.  
Among the remaining features, energy-related features such as coal and natural gas prices, as well as financial indicators like the Euro Stoxx 50 index and EUR/USD exchange rate, also exhibit non-negligible contributions.  
In contrast, macroeconomic uncertainty indices and geopolitical risk show very limited explanatory power, with importance scores close to zero.  
This analysis highlights that policy factors overwhelmingly dominate the drivers of carbon price dynamics, while energy markets and selected financial indicators play secondary roles.

\begin{figure}[H]
  \centering
  \includegraphics[width=9cm]{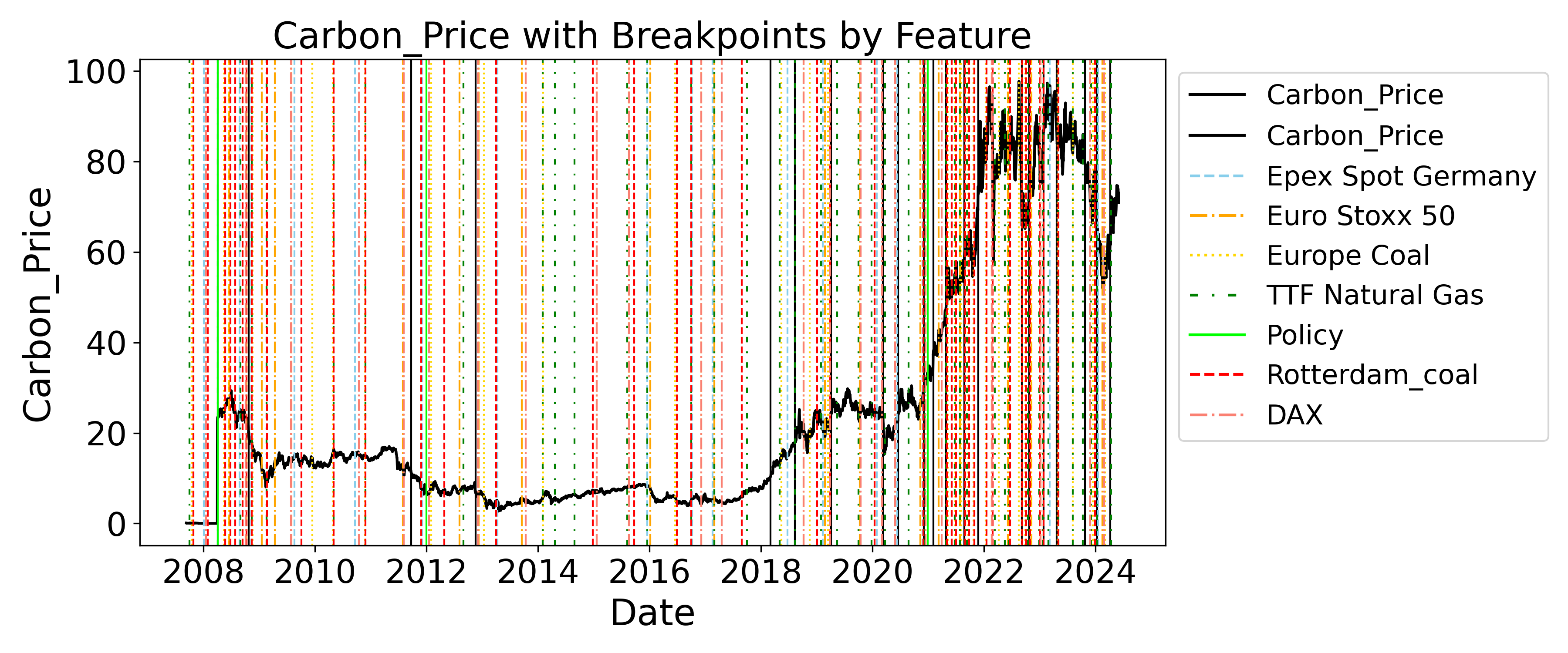}
  \caption{Carbon price time series with detected structural breakpoints by feature}
  \label{fig:timeseries}
\end{figure}

\noindent \textbf{Breakpoint Analysis}.
As shown in Figure \ref{fig:timeseries}, the carbon price time series can be divided into three distinct phases. The first phase, from September 2007 to the end of 2012, is characterized by relatively small price fluctuations, with carbon prices being significantly influenced by policy and coal price disturbances. The second phase, from 2013 to 2019, reflects a relatively stable market, though structural shifts still occurred due to localized impacts from energy prices and policy signals. The third phase, from 2020 to 2024, is marked by a rapid increase and high volatility in carbon prices, mainly driven by the combined influence of natural gas prices, electricity market dynamics, and major policy announcements. In Figure~\ref{fig:timeseries}, the solid black line represents the carbon price, while the dashed lines indicate structural breakpoints associated with various feature features. Notably, breakpoints for Epex Spot Germany, Euro Stoxx 50, TTF natural gas, and policy features appear densely after 2021, clearly highlighting the nonlinear, dynamic, and highly uncertain nature of carbon prices.

Around 2008, due to underdeveloped market mechanisms ~\citep{EcoAct2022}, carbon prices dropped rapidly to near zero, indicating a clear structural break. The carbon price series exhibits positive skewness and high kurtosis, further demonstrating its significant nonnormality.
To enhance the model's ability to learn from the carbon price series and improve prediction accuracy, the dataset is divided into training and testing sets. Common split strategies include “80\%/20\%” or “90\%/10\%.” This study adopts the “80\%/20\%” split, with 80\% of the data used for model training and 20\% for model testing, in order to evaluate the model’s predictive robustness.

Based on the above test results, we also attempt to interpret the structural breakpoints observed in carbon prices. In the first phase (September 2007 – end of 2012), the carbon market was still in its exploratory stage. Issues such as excessive allocation of allowances and frequent economic fluctuations led to generally low and stable carbon prices, with significant influence from policy missteps and coal price disturbances.
(1) In 2007, toward the end of Phase I of the EU ETS (2005–2007), carbon prices plummeted from around €1.5/ton to nearly €0/ton due to the surplus of allowances and the inability to carry over unused allowances into the next phase.
(2) In January 2008, Phase II of the EU ETS officially began. Due to a reduction in the new round of allowance allocations and the partial acceptance of international credits (such as CDM), carbon prices quickly rebounded from nearly €0/ton to around €20/ton.
(3) In September 2008, the global financial crisis severely impacted industrial production in Europe, causing a sharp drop in the demand for carbon allowances. As a result, carbon prices fell rapidly from around €20/ton to approximately €8/ton.
(4) In December 2009, the United Nations Climate Change Conference in Copenhagen failed to reach a legally binding global emissions reduction agreement, undermining market confidence in future carbon prices and causing them to fall from around €15/ton to €12/ton.
(5) In the middle of 2012, the ongoing Eurozone debt crisis continued to suppress industrial activity. Combined with the persistent issue of allowance oversupply, carbon prices dropped to approximately €6.76/ton in June 2012.

Entering the second phase (2013–2019), as reform measures. Such as the Market Stability Reserve (MSR) were gradually implemented, market confidence began to recover \citep{EuropeanCommissionCOP15}, and carbon prices rose significantly. However, localized disruptions caused by energy price fluctuations and policy adjustments still persisted.
(1) In April 2013, the European Parliament rejected the “backloading” proposal aimed at temporarily removing surplus allowances from the market. This triggered disappointment among market participants, leading to a carbon price drop from €7.10/ton to €2.75/ton.
(2) In July 2015, the EU officially adopted the proposal to establish the Market Stability Reserve (MSR) to absorb surplus allowances and improve price flexibility. Although there was little short-term change in carbon prices, market confidence was strengthened, laying the groundwork for future price increases.
(3) Throughout 2018, with the adoption of the reform after 2020 package—which included accelerating the annual reduction of allowances (raising the Linear Reduction Factor to 2.2\%) and tightening the free allocation mechanism—carbon prices surged from €7/ton at the beginning of the year to €25/ton by end of the year, marking an increase of over 250\%.

In the third phase (2020–2024), carbon prices exhibited a combination of rapid increases and high volatility. This period was driven by multiple factors, including the natural gas crisis, the Russia-Ukraine conflict, and new climate policies such as the “Fit for 55” package. The carbon market entered a new stage characterized by both policy driven momentum and structural changes in the energy sector \citep{fan2025cbam}.
(1) In July 2021, the European Union proposed the “Fit for 55” climate policy package, raising the 2030 emissions reduction target from 40\% to 62\% (compared to 2005 levels). This led to a surge in carbon prices from €33/ton at the beginning of the year to €60/ton~\citep{EP2022}.
(2) From February to August 2022, the Russia-Ukraine war caused a spike in natural gas prices, prompting a shift back to coal fired power generation and increasing demand for carbon allowances. As a result, carbon prices reached a historical peak of €97/ton in August 2022.
(3) On February 21, 2023, the EU ETS carbon price surpassed €100/ton for the first time, reaching €101/ton. This was driven by a combination of policy expectations, energy shortages, and growing participation from financial institutions.
(4) From January to February 2024, due to milder winter temperatures, lower electricity demand, and ample renewable energy supply, carbon prices fell from €84/ton to €52/ton, marking the lowest level in 31 months.

\begin{figure}[H]
  \centering
  \includegraphics[width=9cm]{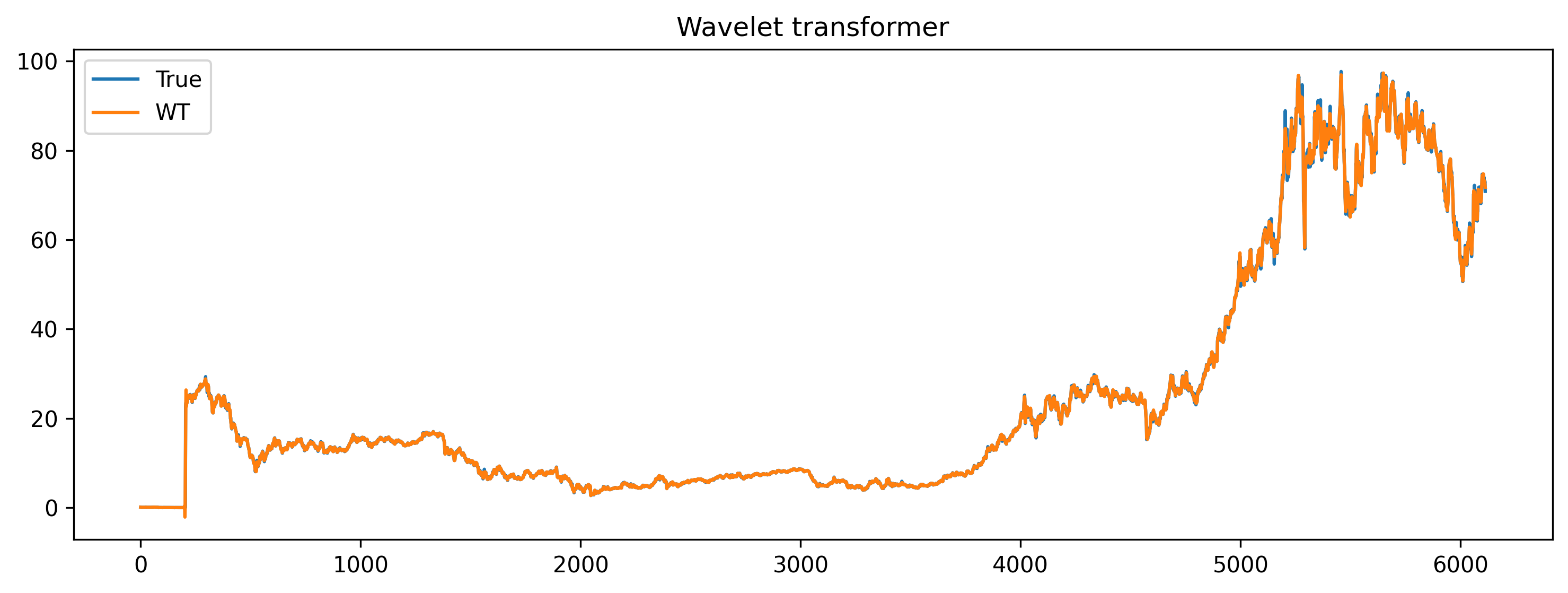}
  \caption{Denoised by Wavelet transformer}
  \label{fig:TCN}
\end{figure}

\noindent \textbf{Wavelet Data Denoising for Carbon Price Series.}
In the original time series, there can exist some notable noises, which can interfere with the ability of machine learning models to learn the intrinsic features of the data. Therefore, it is necessary to remove noise from the original carbon price series before modeling.
To address this, this study applies Wavelet Transform (WT) to process the carbon price time series, decomposing it into approximation components (low frequency trends) and detail components (high-frequency noise). We retain only the approximation component as input to the model, thereby achieving the goal of denoising. A single level decomposition is chosen because excessive decomposition may lead to information loss, which could negatively affect the model's predictive accuracy.
 As shown in Figure~\ref{fig:TCN}, the processed carbon price curve (in orange) is noticeably smoother compared to the original curve (in blue), effectively eliminating abrupt changes and spikes caused by high-frequency fluctuations, while preserving the main trend of price movements.
One key motivation for using WT in this study is to introduce structural breakpoint information on top of the denoised series, in order to explore whether such information can further improve the accuracy of carbon price prediction models under a cleaner signal background.

 \subsection{Comparison of Deep Learning Models}\label{sec:expdlmodel}
 
This study evaluates several deep learning models for carbon price forecasting. 
The compared methods are as follows:

\begin{itemize}
    \item \textbf{BP\&ICSS-WT-LSTM}: Baseline approach that combines Bai–Perron and ICSS structural break detection with wavelet-based LSTM modeling~\citep{lin2022forecasting}.
    \item \textbf{PELT-WT-LSTM (Univariate)}: LSTM model with PELT structural break detection and wavelet transform applied to a single input feature (carbon price).
    \item \textbf{PELT-WT-LSTM (Multivariate)}: Extension of the above model incorporating multiple external influencing factors.
    \item \textbf{PELT-WT-GRU}: GRU-based model with PELT structural break detection and wavelet transform.
    \item \textbf{PELT-WT-TCN}: Temporal Convolutional Network (TCN) model with PELT structural break detection and wavelet transform.
\end{itemize}
To ensure consistency and comparability in the training process of deep learning models for carbon price forecasting, this study adopts a unified training configuration for all models. The Adam optimizer is employed during training due to its adaptive learning rate mechanism, which is well-suited for handling nonstationary time series data. The initial learning
rate is set to 0.001, with momentum parameters 1 and 2 set to
0.9 and 0.999, respectively, to balance training speed and stability. The models are trained with a batch size of 64 for a maximum of 50 epochs. An early stopping mechanism is introduced to prevent overfitting, where training is halted if the validation mean squared error (MSE) does not improve for 10 consecutive epochs. Additionally, 10\% of the training data is allocated as a validation set to continuously monitor model performance. A sliding window approach is used to construct the input data, with each input window consisting of 30 time steps and a stride of 1, ensuring sufficient extraction and modeling of temporal features. Both LSTM and GRU models were built with two hidden layers, each containing 128 units, and a dropout rate of 0.2 was applied between layers to mitigate overfitting. The TCN model was constructed with four residual blocks, each having 64 channels.

\begin{figure}[H]
  \centering
  \includegraphics[width=9cm]{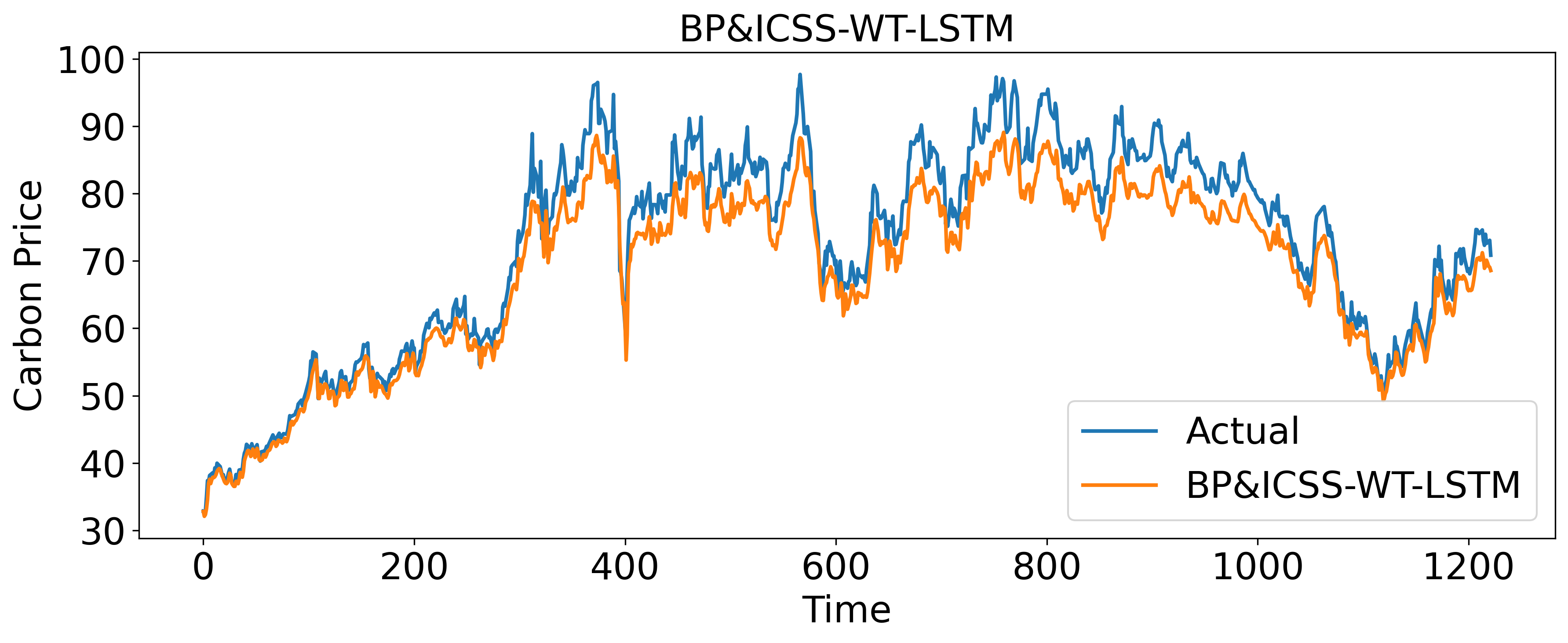}
  \caption{Comparison of actual and predicted carbon prices using the BP\&ICSS-WT-LSTM model}
  \label{fig:lstm}
\end{figure}

\smallskip
\noindent \textbf{BP\&ICSS-WT-LSTM Performance}.
Following the approach of Lin and Zhang~\citep{lin2022forecasting}, Figure~\ref{fig:lstm} presents the prediction results of the BP\&ICSS-WT-LSTM model, reproduced from the baseline study using our updated carbon price dataset. In this configuration, structural breakpoints were detected using a combination of the Bai \& Perron and ICSS methods~\citep{lin2022forecasting}. As a result, the model remains exposed to high-frequency noise and abrupt fluctuations in the time series.
Although the inclusion of breakpoint information helps the model capture shifts in price regimes, BP\&ICSS-WT-LSTM is only able to approximate the overall trend of carbon prices. The model performs poorly around local peaks and troughs, exhibiting amplitude compression, a phenomenon indicating a degree of underfitting. This limitation is particularly evident during periods of high volatility (e.g., around time points 400 and 900), where prediction errors increase significantly.

\begin{figure}[H]
  \centering
  \includegraphics[width=9cm]{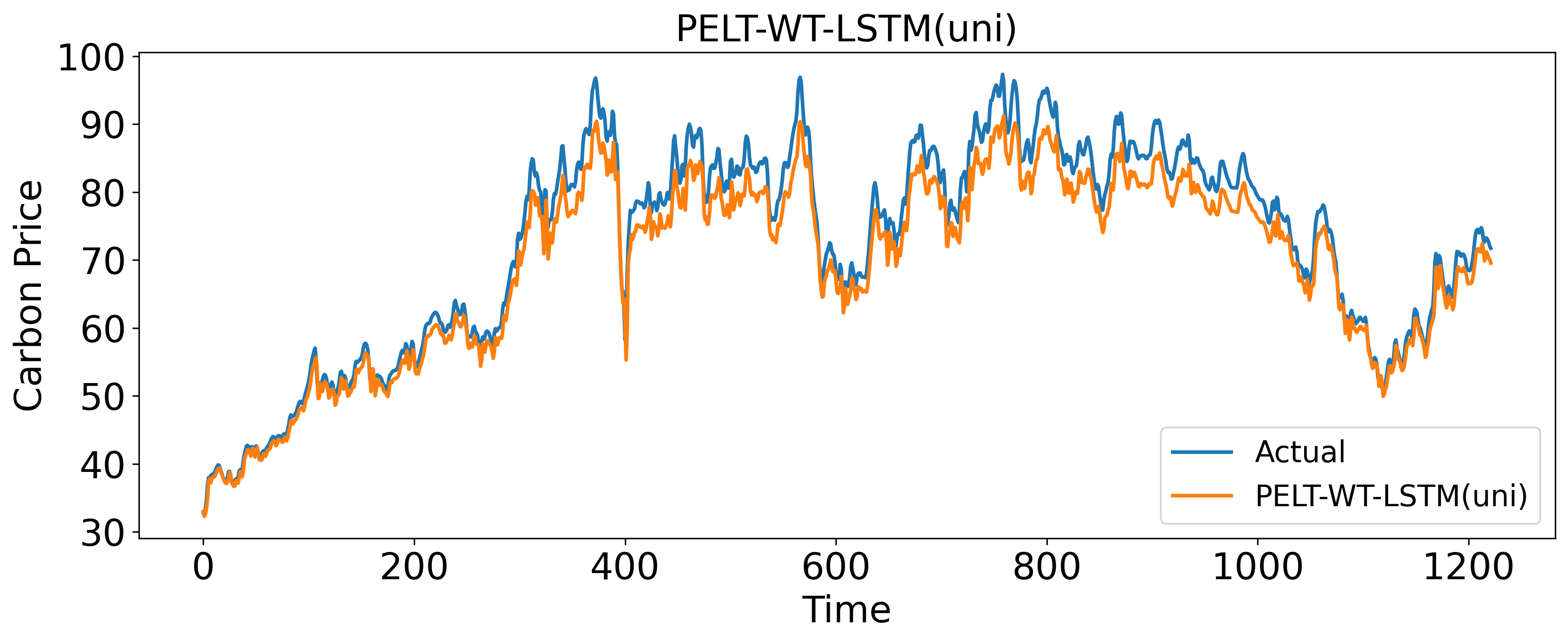}
  \caption{Performance of the PELT-WT-LSTM(uni) model in forecasting carbon prices}
  \label{fig:unilstm}
\end{figure}

\begin{figure}[H]
  \centering
  \includegraphics[width=9cm]{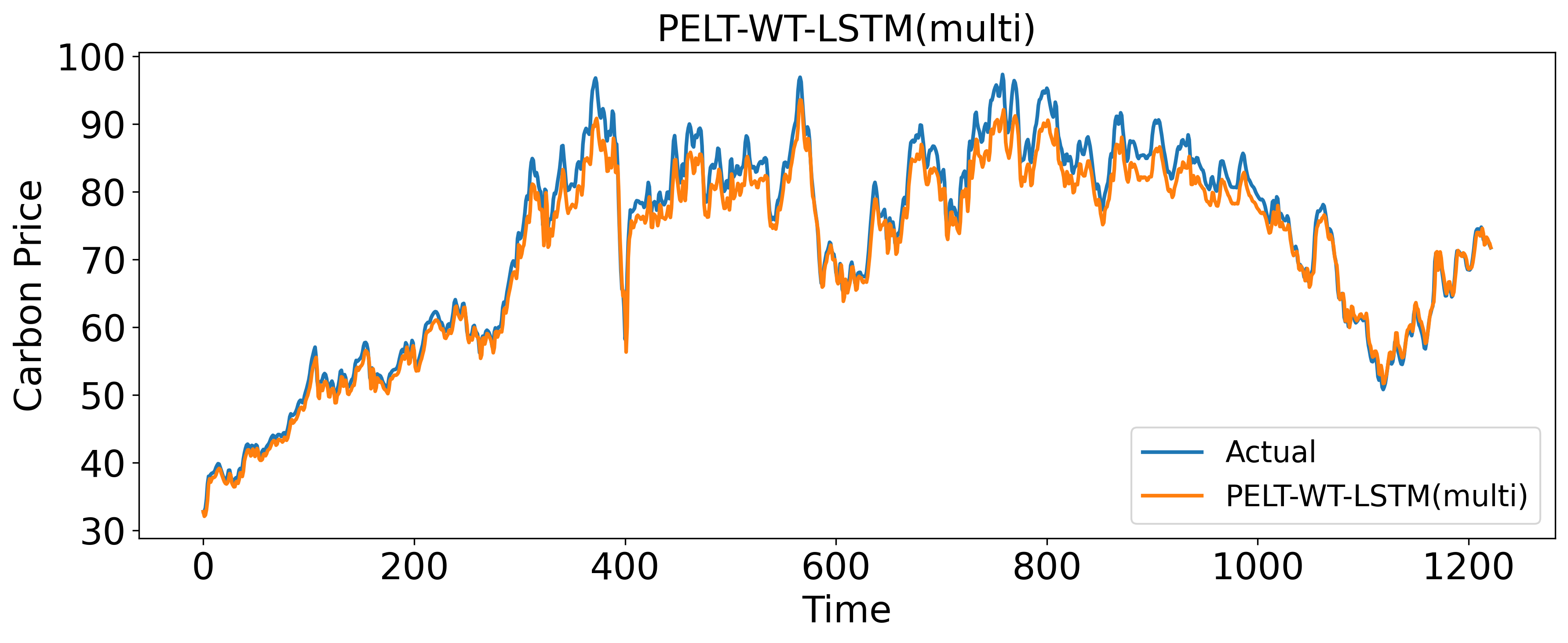}
  \caption{Comparison of actual and PELT-WT-LSTM(multi)}
  \label{fig:comlstm}
\end{figure}

\smallskip
\noindent \textbf{PELT-WT-LSTM (Univariate vs. Multivariate)}.
Figures \ref{fig:unilstm} and Figures \ref{fig:comlstm} reveal the effect of multivariate inputs. The PELT-WT-LSTM(univariate) shows lagged response and error fluctuations. The multivariate version better aligns with actual values in several segments (e.g., step 500 to 900), suggesting enhanced dynamic awareness and forecasting insights.

\begin{figure}[H]
  \centering
  \includegraphics[width=9cm]{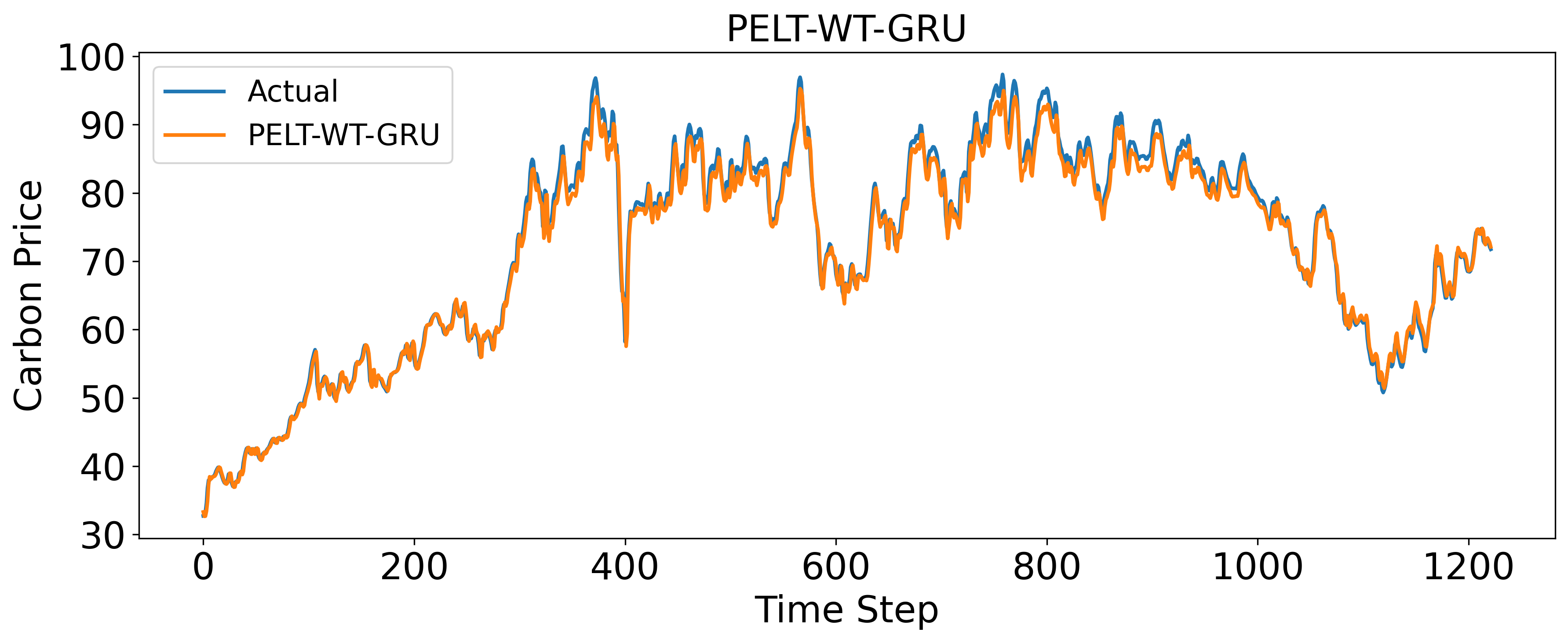}
  \caption{Comparison of actual and PELT-WT-GRU predicted carbon prices}
  \label{fig:comgru}
\end{figure}

\smallskip
\noindent \textbf{PELT-WT-GRU Performance}.
As shown in Figure \ref{fig:comgru}, the GRU model offers structural simplicity with solid fitting capabilities. Prediction trends closely match actual values, especially during price peaks and drops (e.g., step 600–1000). GRU demonstrates a balance between accuracy and training efficiency.

\begin{figure}[H]
  \centering
  \includegraphics[width=9cm]{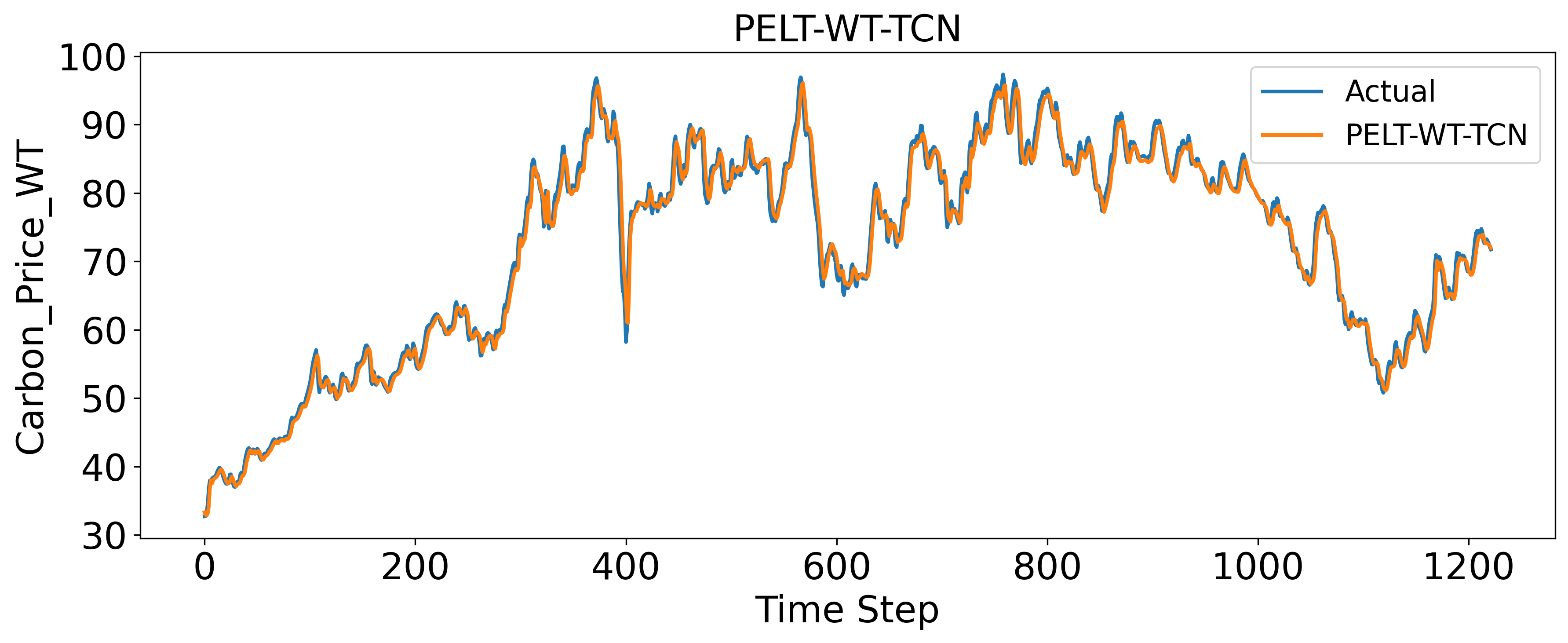}
  \caption{PELT-WT-TCN model prediction performance for carbon price forecasting}
  \label{fig:tcnpred}
\end{figure}

\smallskip
\noindent \textbf{PELT-WT-TCN Performance}.
Figure \ref{fig:tcnpred} indicates that TCN outperforms other models. Its predictions closely follow real values even during sharp fluctuations, thanks to its one dimensional convolution architecture that excels in capturing local temporal features and modelling nonlinearities. 

\subsection{Quantitative Performance Evaluation}\label{sec:expperformance}
 The experiment employed five key performance metrics to quantify model effectiveness, including Mean Absolute Error (MAE), Root Mean Squared Error (RMSE), Mean Absolute Percentage Error (MAPE), the coefficient of determination (R²), and training time. MAE measures the average deviation between predicted and actual values, with smaller values indicating higher prediction accuracy. RMSE emphasizes large errors through squared differences and reflects the overall dispersion of the prediction errors. MAPE expresses the prediction error as a percentage, making it easier to compare across models or datasets. R² evaluates how well the model explains the variance of the target variable, with values closer to 1 indicating better fit. Training time reflects the computational efficiency of each model.

\begin{figure}[H]
  \centering
  \includegraphics[width=9cm]{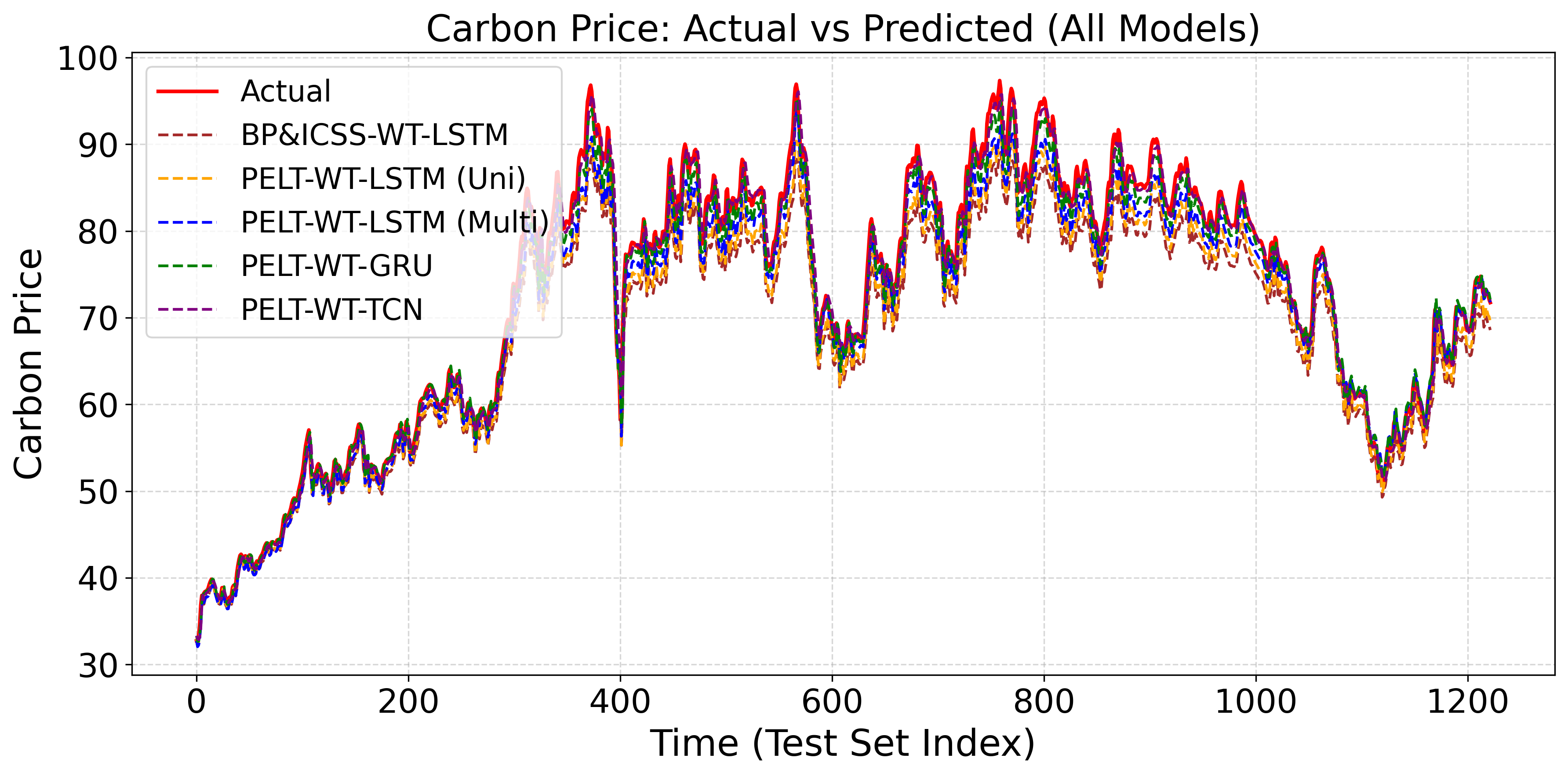}
  \caption{Performance comparison of multiple models in carbon price forecasting}
  \label{fig:compmulti}
\end{figure}

\begin{figure}[H]
  \centering
  \includegraphics[width=9cm]{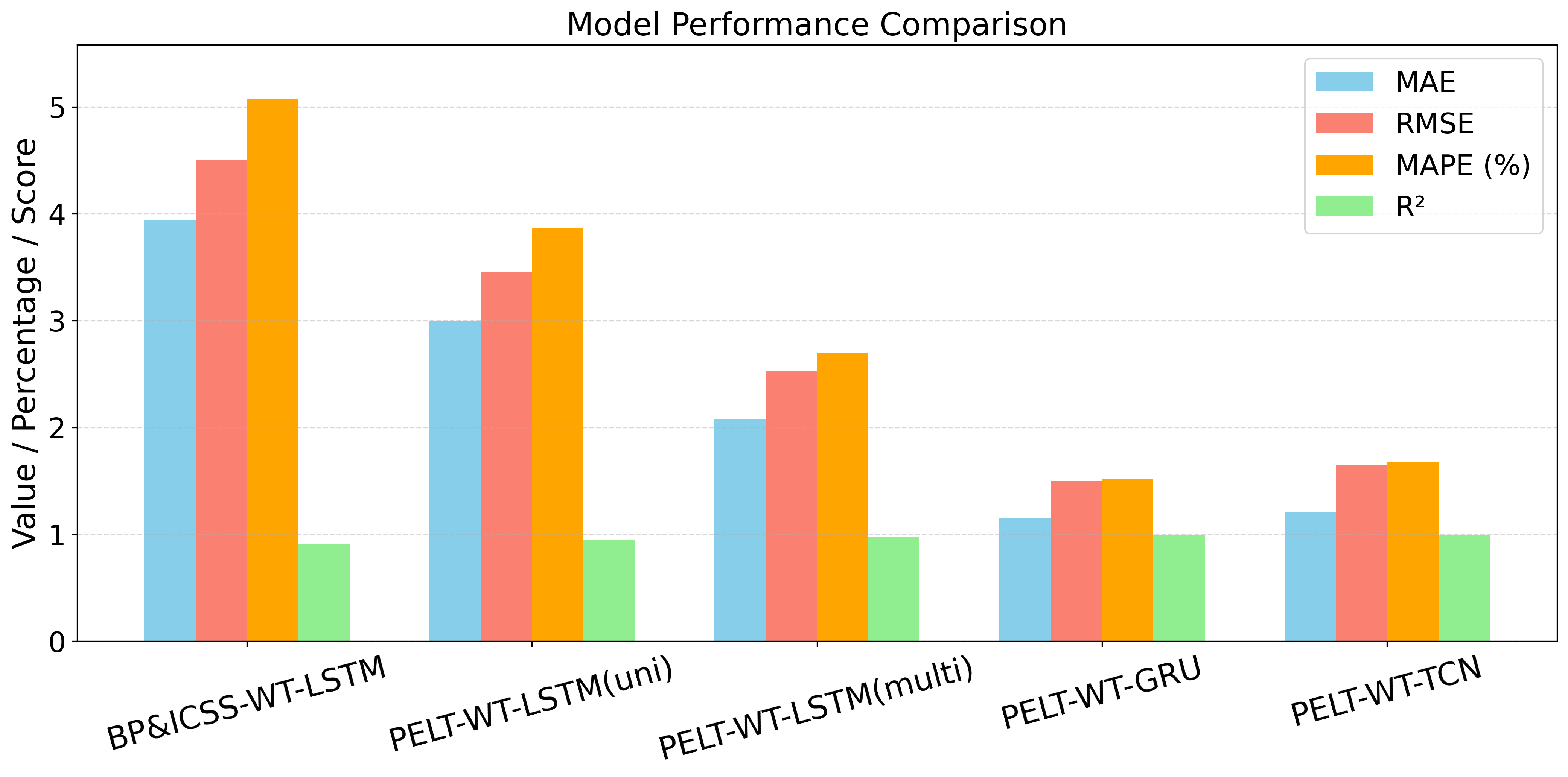}
  \caption{Comparison of model performance metrics}
  \label{fig:compmet}
\end{figure}

Table \ref{tab:model-performance}, Figure \ref{fig:compmulti} and Figure \ref{fig:compmet} present five key performance metrics—Mean Absolute Error (MAE), Root Mean Squared Error (RMSE), Mean Absolute Percentage Error (MAPE), R² (coefficient of determination), and training time—for various deep learning models used in time series forecasting. The results reveal clear differences in accuracy and computational efficiency among the models:

\begin{table}[H]
\centering
\caption{Performance comparison for carbon price prediction}
\label{tab:model-performance}
\resizebox{0.7\textwidth}{!}{
\begin{tabular}{l|c|c|c|c}
\hline
\textbf{Model} & \textbf{MAE} & \textbf{RMSE} & \textbf{MAPE (\%)} & \textbf{$R^2$}  \\
\hline
BP\&ICSS-WT-LSTM & 4.6345 & 5.3878 & 5.8731 & 0.8712  \\
PELT-WT-LSTM (uni) & 2.3627 & 2.7488 & 3.0582 & 0.9664  \\
PELT-WT-LSTM (multi) & 1.8192 & 2.2967 & 2.3267 & 0.9765 \\
PELT-WT-GRU & 1.3308 & 1.6987 & 1.7401 & 0.9872  \\
PELT-WT-TCN & \textbf{1.1855} & \textbf{1.5866} & \textbf{1.6451} & \textbf{0.9888}  \\
\hline
\end{tabular}
}
\end{table}


\smallskip
\noindent \textbf{PELT-WT-TCN}.
The TCN model delivers the best overall performance, achieving the lowest error metrics (MAE = 1.1855, RMSE = 1.5866, MAPE = 1.6451\%), indicating minimal deviation between predicted and actual values. It also achieves the highest R² score of 0.9888, meaning it explains nearly all variance in the data. However, this comes at the cost of the longest training time (48.7 seconds), likely due to its deeper architecture or complex convolutional operations.

\smallskip
\noindent \textbf{PELT-WT-GRU}.
The GRU model strikes a strong balance between performance and speed. With MAE = 1.3308, RMSE = 1.6987, MAPE = 1.7401\%, it ranks closely behind TCN in accuracy. Importantly, it has the fastest training time (19.6 seconds), making it highly suitable for time sensitive applications. Its R² of 0.9872 also confirms excellent predictive reliability.

\smallskip
\noindent \textbf{PELT-WT-LSTM(Multivariate)}.
The PELT-WT-LSTM(Multivariate) model, which incorporates multiple input features, performs well in terms of prediction accuracy (MAE = 1.8192, RMSE = 2.2967, MAPE = 2.3267\%) with a respectable R² of 0.9765. The training time is moderate (24.9 seconds), making it a viable option when balancing accuracy and computational cost.

\smallskip
\noindent \textbf{PELT-WT-LSTM(Univariate)}.
This model uses only a single input feature and underperforms compared to its multivariate counterpart. With MAE = 2.3627, RMSE = 2.7488, MAPE = 3.0582\%, and R² = 0.9664, it shows limited predictive power. The training time is around 24.7 seconds, similar to the multivariate version.

\smallskip
\noindent \textbf{BP\&ICSS-WT-LSTM}.
The BP\&ICSS-WT-LSTM configuration performs the worst overall, with high error rates (MAE = 4.6345, RMSE = 5.3878, MAPE = 5.8731\%) and the lowest R² score (0.8712), suggesting a poor fit to the data. Despite its moderate training time (24.5 seconds), its low accuracy renders it unsuitable for practical applications.

While PELT-WT-TCN achieves the highest accuracy, it requires longer training time. PELT-WT-GRU provides a approximately optimal balance between speed and performance, making it ideal for use in real time. PELT-WT-LSTM(Multi) also performs well with multiple features. In contrast, PELT-WT-LSTM(Uni) and BP\&ICSS-WT-LSTM models fall short, particularly the latter, which is not recommended for deployment due to its poor performance.
\subsection{Scalability and Residual Distribution Analysis}\label{sec:expscalabilityresidual}
\begin{figure}[H]
\centering
\includegraphics[width=9cm]{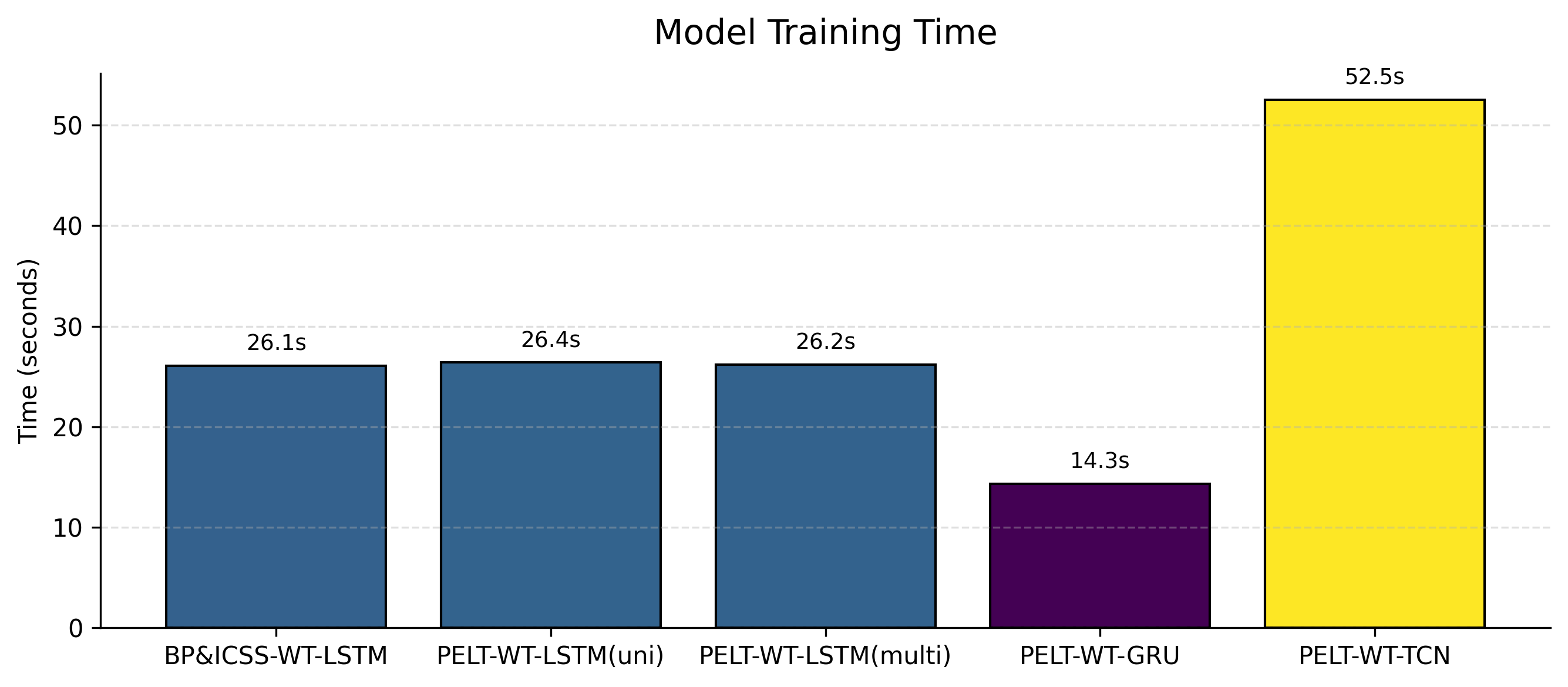}
\caption{Model training times}
\label{fig:tim}
\end{figure}
Figure \ref{fig:tim} provides a detailed comparison of the training times required by different models under identical training conditions, highlighting significant differences in computational resource consumption. Overall, the PELT-WT-GRU model stands out with the fastest training time of just 14.3 seconds, making it particularly advantageous in scenarios that are sensitive to time or constrained by resources. In contrast, the BP\&ICSS-WT-LSTM, PELT-WT-LSTM(univariate), and PELT-WT-LSTM(Multivariate) show very similar training durations, 26.1, 26.4, and 26.2 seconds respectively, indicating that the PELT-WT-LSTM architecture maintains relatively stable training efficiency regardless of input dimensionality.  Although their training times are moderate, these LSTM-based models are slightly less efficient than PELT-WT-GRU. On the other hand, the PELT-WT-TCN model requires the longest training time, reaching 52.5 seconds, nearly 3.5 times that of PELT-WT-GRU. This extended duration is primarily due to the TCN’s deep convolutional structure and multiple stacked layers, which make the training process more computationally intensive and complex.

\begin{figure}[H]
  \centering
  \includegraphics[width=9cm]{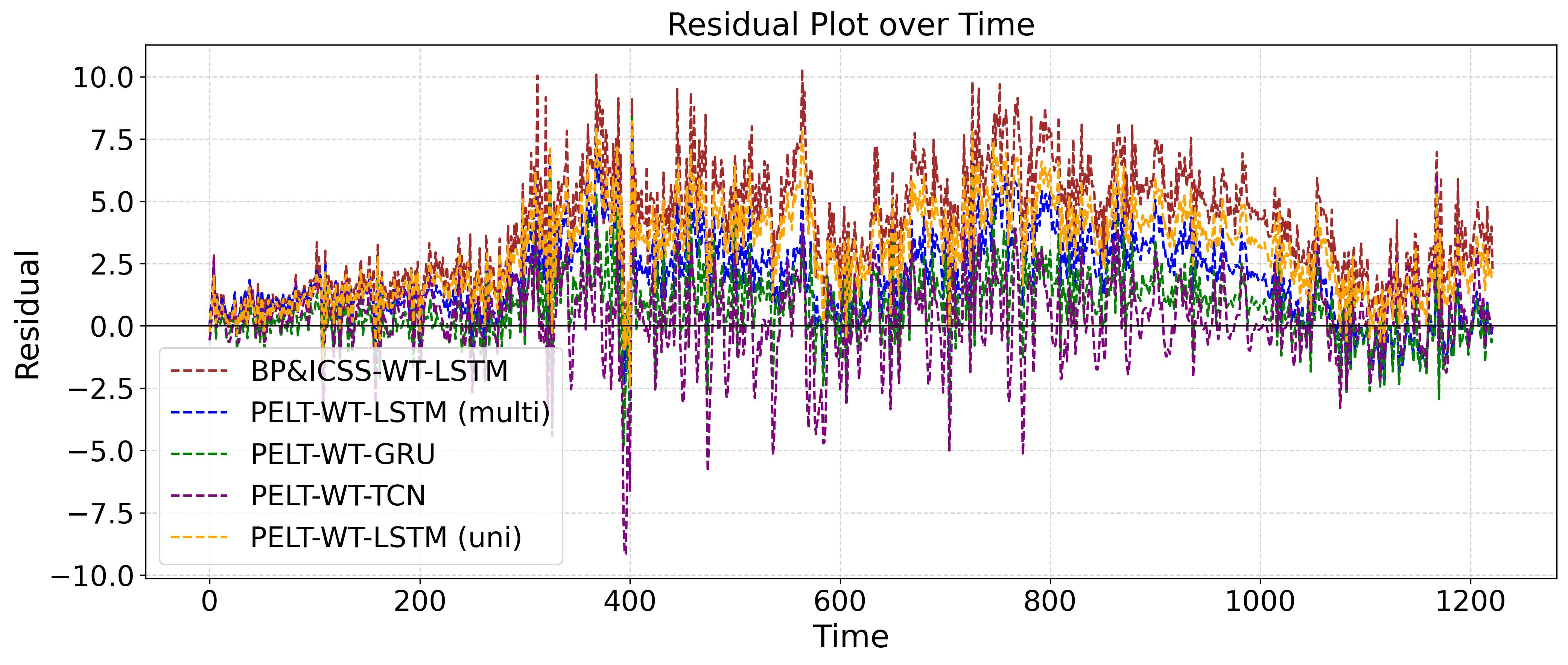}
  \caption{Residual distributions across time}
  \label{fig:residi}
\end{figure}
Figure~\ref{fig:residi} illustrates the temporal distribution of residuals (actual minus predicted values) for five deep learning models. The PELT-WT-TCN model demonstrates relatively constrained residual variation, with values fluctuating closely around zero, which may reflect its strength in capturing long range temporal dependencies through dilated causal convolutions \citep{bai2018empirical}. At the other end of the spectrum, the BP\&ICSS-WT-LSTM exhibits broader and more irregular residual patterns, with multiple spikes exceeding ±10, potentially indicating heightened sensitivity to input volatility or sequence length. PELT-WT-GRU and PELT-WT-LSTM (Multivariate) present moderate residual variability, while PELT-WT-LSTM (Univariate) maintains a more symmetric error pattern with smaller amplitude, although an increased concentration of negative deviations is observed in the latter part of the series. These findings highlight the extent to which architectural design shapes residual dynamics and suggest that residual stability may serve as a practical proxy for model generalization in time series forecasting.

\begin{figure}[H]
  \centering
  \includegraphics[width=9cm]{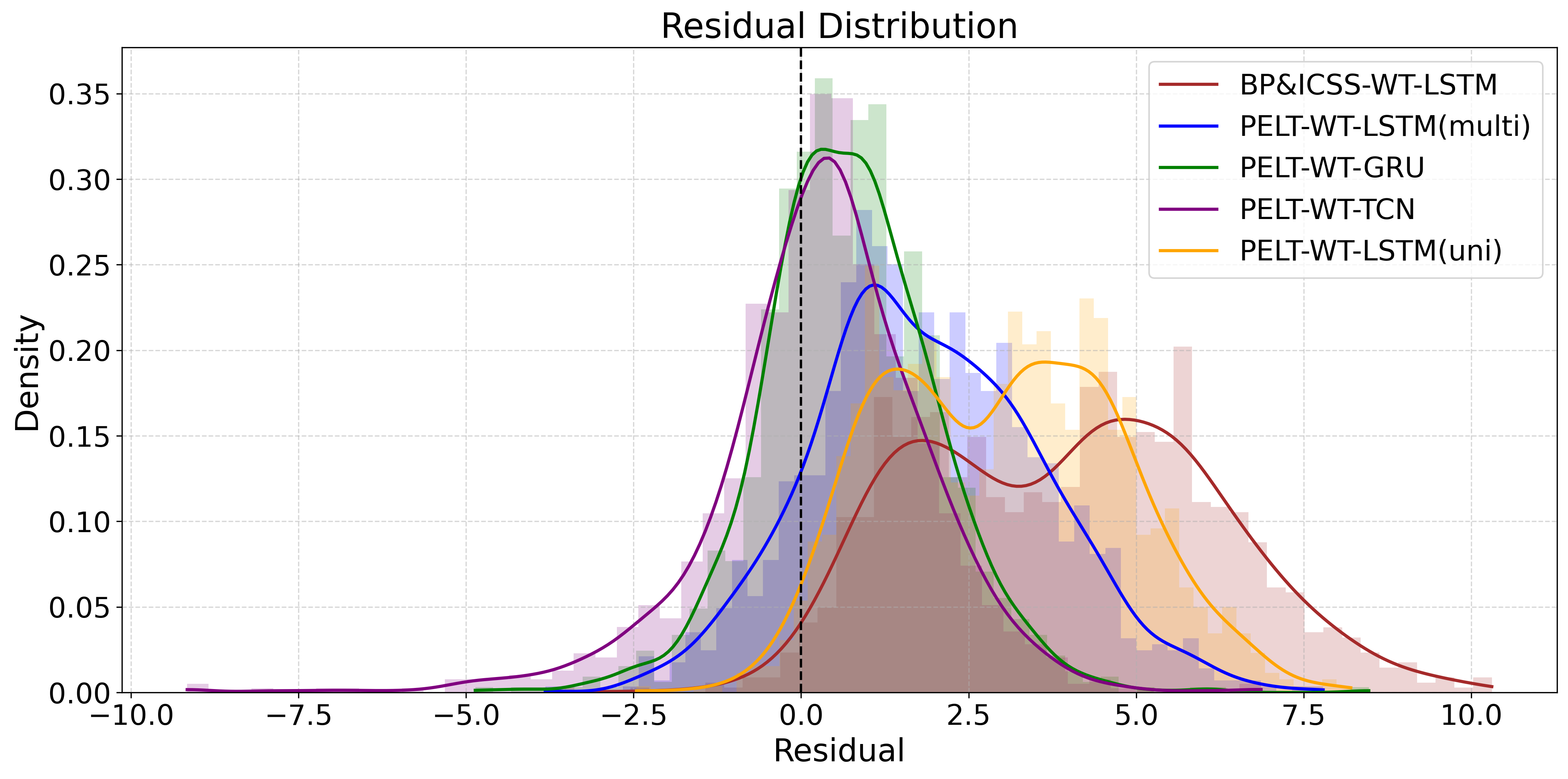}
  \caption{Distribution of residuals density}
  \label{fig:dires}
\end{figure}
Figure~\ref{fig:dires} illustrates the residual distributions of different models. Among all the methods, the PELT-WT-TCN model exhibits the most concentrated and symmetric density around zero, indicating both low residual variance and minimal bias. The PELT-WT-GRU and PELT-WT-LSTM(Multivariate) models demonstrate moderately peaked curves, with PELT-WT-GRU showing slight right skewness suggestive of occasional overestimation. In addition, the PELT-WT-LSTM(Univariate) distribution is broader and shifted rightward, reflecting a greater frequency of positive errors. The BP\&ICSS-WT-LSTM displays the widest and most asymmetric distribution, characterized by a pronounced right tail, which implies a tendency toward persistent over prediction. The distributional characteristics of residuals provide an additional layer of model evaluation beyond time domain analysis \citep{lawrance1985}, revealing that PELT-WT-TCN and PELT-WT-GRU are more consistent and less prone to systematic error compared to the BP\&ICSS-WT-LSTM.

\section{Conclusion}
\label{sec:Conclusion}

This study proposed a hybrid forecasting framework that integrates structural breakpoint detection, wavelet denoising, and advanced deep learning models to enhance the prediction of EU carbon prices. Breakpoint detection using PELT enables the model to capture regime shifts caused by policy changes and market shocks, while wavelet transform reduces high-frequency noise and improves stability. Together, these preprocessing steps provide denoised and inputs with regime awareness for subsequent forecasting. Experimental results show that incorporating structural and denoised information significantly improves predictive accuracy. Among the tested models, PELT-WT-TCN achieves the lowest error metrics, while PELT-WT-GRU demonstrates a favorable balance between efficiency and performance, making it more suitable for real time applications. The comparison confirms that combining structural awareness with multiscale learning is essential for improving robustness and interpretability in carbon price forecasting.
Overall, the findings contribute to the growing literature on hybrid modeling for financial and energy markets, demonstrating the benefits of integrating advanced time series decomposition with deep learning. Future research may focus on improving model interpretability and extending the dataset with more indicators related to policy and macroeconomic factors.

\section*{Declaration of Competing Interest}
The authors declare that they have no known competing financial interests or personal relationships that could have appeared to influence the work reported in this paper.



\bibliographystyle{elsarticle-harv} 

\bibliography{cas-refs}

\end{document}